\documentclass[10pt,twocolumn,letterpaper]{article}

\usepackage[pagenumbers]{style}

\usepackage{graphicx}
\usepackage{amsmath}
\usepackage{amssymb}
\usepackage{booktabs}

\usepackage{multicol}
\usepackage{multirow}
\usepackage{array}
\newcolumntype{P}[1]{>{\centering\arraybackslash}p{#1}}

\usepackage{wrapfig}

\usepackage{cprotect}
\usepackage{booktabs}

\usepackage[square,sort,comma,numbers]{natbib}
\usepackage{multibib}
\newcites{appendix}{Appendix References}

\usepackage[pagebackref,breaklinks,colorlinks]{hyperref}

\usepackage{marvosym}

\usepackage[capitalize]{cleveref}
\crefname{section}{Sec.}{Secs.}
\Crefname{section}{Section}{Sections}
\Crefname{table}{Table}{Tables}
\crefname{table}{Tab.}{Tabs.}

\usepackage[normalem]{ulem} 
\def\Name{Uni-Perceiver}

\begin{document}

\title{Uni-Perceiver: Pre-training Unified Architecture for Generic Perception for Zero-shot and Few-shot Tasks}

\author{
    Xizhou Zhu$^{1*}$, 
    Jinguo Zhu$^{2*\dag}$, 
    Hao Li$^{3*\dag}$, 
    Xiaoshi Wu$^{3*\dag}$
\vspace{0.2em}\\
    Xiaogang Wang$^{3}$, 
    Hongsheng Li$^{3}$, 
    Xiaohua Wang$^{2}$, 
    Jifeng Dai$^{1}$\textsuperscript{\Letter}
\vspace{0.4em}\\
    $^{1}$SenseTime Research \quad $^{2}$Xi'an Jiaotong University \\
    $^{3}$CUHK-SenseTime Joint Laboratory, The Chinese University of Hong Kong \\ 
    \texttt{\small \{zhuwalter,\,daijifeng\}@sensetime.com}
\\
    \texttt{\small lechatelia@stu.xjtu.edu.cn},\ \ \texttt{\small \{haoli,\,wuxiaoshi\}@link.cuhk.edu.hk}
\\
    \texttt{\small \{xgwang,\,hsli\}@ee.cuhk.edu.hk},\ \ \texttt{\small xhw@mail.xjtu.edu.cn}
}
\maketitle

\thispagestyle{empty}

\footnotetext[1]{\noindent Equal contribution. $^{\dag}$This work is done when Jinguo Zhu, Hao Li, and Xiaoshi Wu are interns at SenseTime Research. \textsuperscript{\Letter}Corresponding author.}

\begin{abstract}
Biological intelligence systems of animals perceive the world by integrating information in different modalities and processing simultaneously for various tasks. In contrast, current machine learning research follows a task-specific paradigm, leading to inefficient collaboration between tasks and high marginal costs of developing perception models for new tasks. In this paper, we present a generic perception architecture named Uni-Perceiver, which processes a variety of modalities and tasks with unified modeling and shared parameters. Specifically, Uni-Perceiver encodes different task inputs and targets from arbitrary modalities into a unified representation space with a modality-agnostic Transformer encoder and lightweight modality-specific tokenizers. Different perception tasks are modeled as the same formulation, that is, finding the maximum likelihood target for each input through the similarity of their representations. The model is pre-trained on several uni-modal and multi-modal tasks, and evaluated on a variety of downstream tasks, including novel tasks that did not appear in the pre-training stage. Results show that our pre-trained model without any tuning can achieve reasonable performance even on novel tasks. The performance can be improved to a level close to state-of-the-art methods by conducting prompt tuning on 1\% of downstream task data. Full-data fine-tuning further delivers results on par with or better than state-of-the-art results. Code shall be released.

\end{abstract}

\pagebreak

\section{Introduction}
\label{sec:intro}
\begin{figure}[t]
\centering
\includegraphics[width=0.48\textwidth]{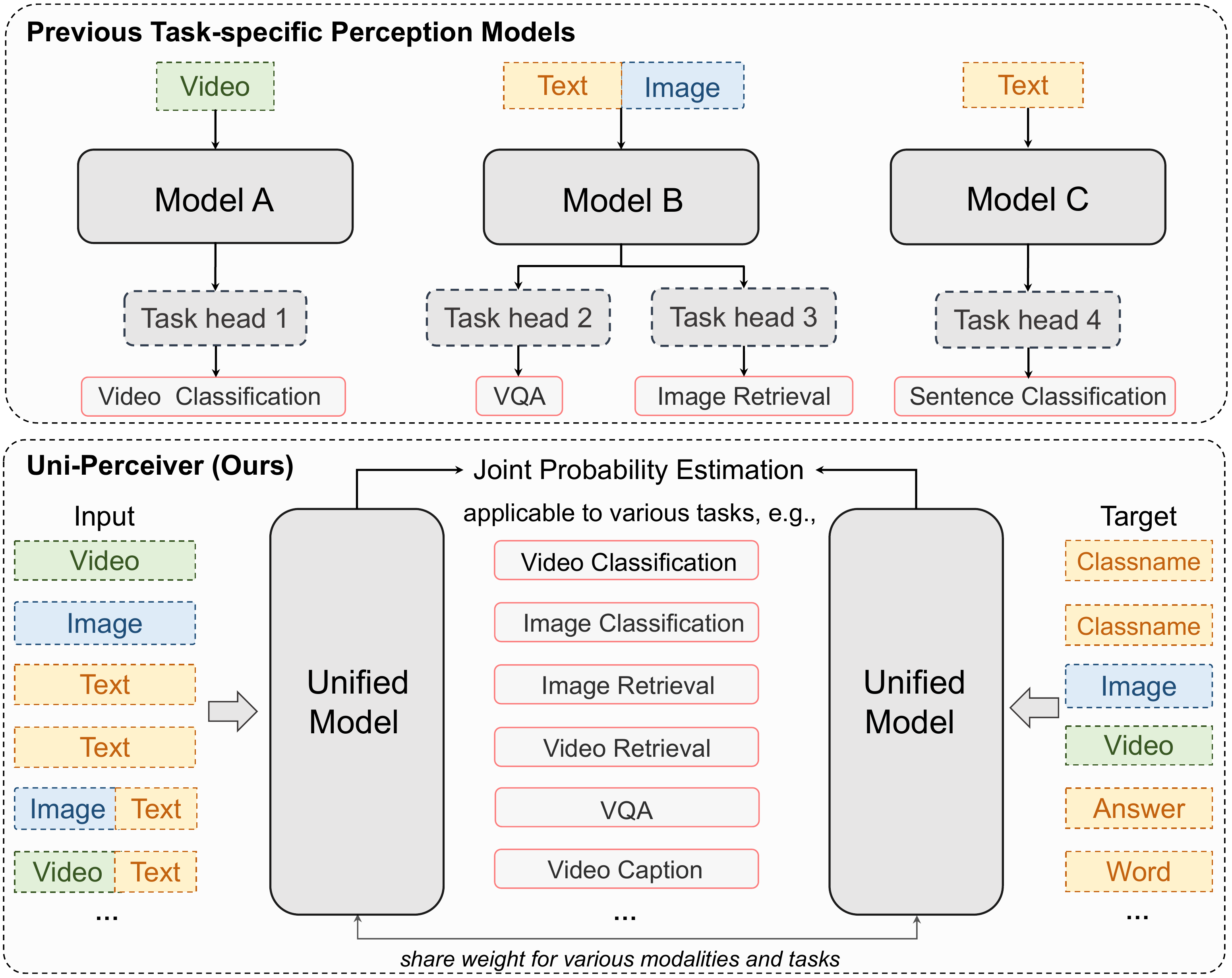}
\caption{Comparing previous task-specific perception models with our proposed \Name, which processes various modalities and tasks with a single siamese model and shared parameters.}
\label{fig:comparison}
\vspace{-0.5em}
\end{figure}

Biological intelligence systems of animals perceive the world by receiving information in different modalities, integrating with the complex central nervous system, and processing simultaneously for different tasks.
However, designing a generic artificial perception model that handles multiple modalities and numerous tasks has always been considered too difficult.
To simplify this problem, previous machine learning research has focused on developing specialized models for inputs from certain restricted modality, \eg, Convolutional Neural Networks~\cite{CNNSURVEY} for visual recognition and Transformers~\cite{vaswani2017attention} for natural language processing.  
Recently, Transformers have been proved to have competitive performance in more scenarios such as image~\cite{dosovitskiy2020image,liu2021Swin,pmlr-v139-touvron21a,wang2021pyramid,Yuan_2021_ICCV,touvron2021cait,chen2021crossvit,zhou2021deepvit} and video~\cite{arnab2021vivit,gberta_2021_ICML,Zhang_2021_ICCV} recognition, which triggers a new paradigm of designing unified architectures for different modalities. 
Following this paradigm, recent works~\cite{kim2021vilt,Radford2021LearningTV,hu2021unit,Akbari2021VATTTF} adopt Transformers as the backbone for multi-modal applications such as visual-linguistic recognition. They convert the inputs from different modalities into unified input token sequences with modality-specific tokenizers. Models are pre-trained with large-scale multi-modal datasets, and then adapted to downstream tasks with fine-tuning.

Despite the ability of processing multi-modal information with unified architectures, current methods still require specific design and training for different tasks.
This limitation is caused by two reasons. 
First, the input of a particular model is the combination of specific modalities required by its target task. 
Second, previous works require prediction heads specifically designed and trained for the target tasks. 

We argue that this task-specific paradigm conflicts with the objective of designing generic perceptual models.
Specifically, during pre-training, the specialised designs for different tasks hinder the collaboration between tasks, which may hurt the representational capacity. 
Meanwhile, when a pre-trained model is applied to a new task, the input format and the prediction head need to be re-designed and fine-tuned on sufficient downstream data. 
Considerable effort in collecting and annotating data is required.
Also, all parameters need to be copied and maintained for each downstream task, which becomes inefficient and inconvenient as the number of tasks and the model size grow.
On the other hand, when fine-tuning is performed with insufficient training data, it may forget the pre-trained knowledge that is beneficial to the downstream task, thereby hurting generalization performance~\cite{chen2019catastrophic}.
All of these issues increase the marginal cost of developing perception models for new tasks and limit the capability to meet the rapidly growing demands of diverse scenarios, indicating task-specific paradigm is not suitable for generic perceptual modeling.

Our core idea is to replace task-specific designs by encoding different task inputs and targets from arbitrary modalities into a unified representation space, and model the joint probability of inputs and targets through the similarity of their representations.
This design eliminates the gap between the formulations of different perception tasks, and therefore encourages the collaboration between different modalities and tasks in representation learning. 
Moreover, by aligning the formulations of pre-training and downstream tasks, the knowledge can be better transferred when applying the pre-trained model to the target tasks. The model can even conduct zero-shot inference on novel tasks that do not appear in the pre-training stage.

In this paper, we propose a unified architecture named \Name, which processes various modalities and tasks with a single siamese model and shared parameters.
Specifically, the task inputs and targets from arbitrary combinations of modalities are first converted into unified token sequences with lightweight modality-specific tokenizers. 
The sequences are then encoded by a modality-agnostic Transformer encoder into a unified representation space.
Different perception tasks are modeled as the same formulation, finding the maximum likelihood target for each input through the similarity of their representations, so as to facilitate the generic perceptual modeling.

\Name\ is pre-trained on various uni-modal tasks such as image / video classification and language modeling, and multi-modal tasks such as image-text retrieval and language modeling with image clues.
When applied to downstream tasks, thanks to the generic modeling of perception tasks, the pre-trained model shows the ability of zero-shot inference on novel tasks that did not appear in the pre-training stage.
Moreover, the performance can be further boosted with additional task-specific data. For the few-shot scenario, we adapt the model to downstream tasks with prompt tuning~\cite{liu2021pre}, where only a small amount of additional parameters are optimized for specific tasks. The performance of our model can be further improved with full-model fine-tuning on sufficient downstream training data.

We pre-train our model on several uni-modal and multi-modal tasks, and evaluate its performance on a variety of downstream tasks, including novel tasks that did not appear in the pre-training stage. Results show that our pre-trained model without any tuning can achieve reasonable performance even on novel tasks. Its performance can be boosted to a level close to state-of-the-art methods by conducting prompt tuning with 1\% of the downstream task data. When fine-tuning the pre-trained model with 100\% of the target data, our model achieves result on par with or better than state-of-the-art methods on almost all the tasks, which demonstrates the strong representation ability.

\section{Related Works}
\label{sec:related}
\noindent\textbf{Architecture.~}
For visual recognition, Convolutional Neural Networks (CNN)~\cite{CNNSURVEY} used to be the main architecture paradigm.
Motivated by the success of Transformers in natural language processing~\cite{vaswani2017attention,Brown2020LanguageMA,devlin2018bert,liu2019roberta, lan2019albert,lewis2019bart}, attempts have been made to apply Transformers to image and video modalities.
For image recognition, vision Transformers~\cite{dosovitskiy2020image,liu2021Swin,pmlr-v139-touvron21a,wang2021pyramid,Yuan_2021_ICCV,touvron2021cait,chen2021crossvit,zhou2021deepvit} replace CNNs by an image patch tokenizer and a transformer encoder, which have been proved to obtain competitive performance as CNNs.
\cite{arnab2021vivit,gberta_2021_ICML,Zhang_2021_ICCV} make attempts to apply Transformers on video recognition in a convolution-free fashion. 
For visual-linguistic recognition, recent works~\cite{Lu2019ViLBERTPT,tan2019lxmert,li2019visualbert,su2019vl,Li2020UnicoderVLAU,chen2020uniter,zhou2020unified,Qi2020ImageBERTCP} also adopt Transformers as the backbone, while they usually take regional features as inputs, which are typically extracted by off-the-shelf object detectors (\eg, Faster R-CNN~\cite{ren2015faster} pre-trained on Visual Genome~\cite{krishna2017visual}). \cite{huang2020pixel} attempts to eliminate the need for object detectors by directly extracting features from the raw pixels with CNNs. 
\cite{kim2021vilt,Radford2021LearningTV,hu2021unit,Akbari2021VATTTF} take a further step by applying Transformers to raw image patches and word tokens.
Transformers have enabled a unified architecture paradigm for different modalities, which only need the modality-specific tokenizers to convert inputs from different modalities into unified input token sequences.

Nevertheless, previous architecture requires prediction heads specifically designed and trained for different perception tasks.
Instead, we replace the task-specific design by encoding different task inputs and targets into a unified representation space, and model their relationship by representational similarity. This modification enables our model to conduct zero-shot inference even on novel downstream tasks that did not appear in the pre-training stage.

\vspace{0.5em}
\noindent\textbf{Pre-training.~}
Large-scale pre-training has achieved great success in the field of deep learning, which can alleviate the data-hungry challenge and improve the performance of downstream tasks~\cite{xu2021pre}.
For image recognition, pre-training is usually performed on image classification datasets, \eg, ImageNet~\cite{deng2009imagenet}. Video recognition networks are either pre-trained on image classification or video classification datasets, \eg, Moments in Time~\cite{monfort2019moments} and Kinetics~\cite{kay2017kinetics}. 
In natural language processing, self-supervised language modeling~\cite{Brown2020LanguageMA,devlin2018bert,liu2019roberta, lan2019albert,lewis2019bart} is adopted for pre-training on large-scale unlabeled corpora~\cite{qiu2020pre}. 
Specifically, GPT~\cite{Brown2020LanguageMA} performs the auto-regressive pre-training, which optimizes the probability of the next word conditioned on previous words. BERT~\cite{devlin2018bert} uses masked language modeling (MLM) and next sentence prediction (NSP) for pre-training.
These pre-trained models can serve as robust feature extractors for downstream tasks with small architecture modifications.

Recent years have witnessed interest in large-scale cross-modal pre-training~\cite{xu2021pre}. Compared with uni-modal pre-training, cross-modal pre-training needs to align information from different modalities.
Such pre-training is usually performed on image-text pairs collected from Internet~\cite{sharma2018conceptual,ordonez2011im2text,yfcc,changpinyo2021cc12m} and manual annotated visual-linguistic datasets~\cite{lin2014microsoft,krishna2017visual,ordonez2011im2text}.
Moreover, various pre-training objectives are also proposed to utilize these datasets effectively.
The most widely used objectives are image-text retrieval~\cite{sun2019videobert, cbt, lu2019vilbert,b2t2,tan2019lxmert,li2019visualbert,clip}, masked language modeling with image clues~\cite{sun2019videobert, cbt, lu2019vilbert,b2t2,tan2019lxmert,li2019visualbert,su2019vl}, and masked region modeling~\cite{cbt, lu2019vilbert,tan2019lxmert,su2019vl,chen2020uniter}. 
Among them, masked region modeling requires regional features extracted by off-the-shelf object detectors.
More recently, CLIP~\cite{clip} has verified the effectiveness of only performing image-text retrieval pre-training on huge webly collected data.

Previous multi-task pre-training requires task-specific heads, which hinders the collaboration among different tasks.
Instead, we encode different task inputs and targets into a unified representation space, and model their relationship by a unified representational similarity, which enables the collaboration between different modalities and tasks.
Our pre-training tasks include image and video classification, language modeling with and without image clues, and image-text retrieval. We do not use regional features and the corresponding pre-training tasks.

\vspace{0.5em}
\noindent\textbf{Prompt Tuning.~}
As an alternative solution to fine-tuning, prompt tuning has recently been proposed in the NLP community, which originated from prompting methods~\cite{liu2021pre}.
In prompting, specially designed natural language tokens, or namely prompts, are inserted into the input sequence as hints for the target tasks. These prompt inputs are used to query a large language model (\eg, GPT-3~\cite{Brown2020LanguageMA}). Methods~\cite{autoprompt:emnlp20,Jiang2020HowCW} have been proposed to automate the prompt engineering process. 
The prompting process does not tune any of the parameters, which is empirically sub-optimal compared to fine-tuning~\cite{liu2021gpt}.

Prompt tuning~\cite{Lester2021ThePO} is proposed to replace hard language prompts with learnable prompt tokens that can be updated through gradient back-propagation, while other parameters are still kept fixed.
Other than adding learnable input tokens, Prefix-Tuning~\cite{li2021prefixtuning} adds learnable prompts to each layer of the Transformer to boost the model capacity.
For few-shot scenario, \cite{gao2021making} proves that prompt tuning can be much better than traditional fine-tuning. When the training data is sufficient, prompt tuning performs slightly worse than fine-tuning~\cite{liu2021gpt}. However, the performance gap from full-model fine-tuning closes up as the pre-trained model gets larger~\cite{Lester2021ThePO,liuPtuningv2}.
Inspired by the success of prompt tuning in NLP, \cite{Zhou2021LearningTP} applies prompt tuning to visual-linguistic pre-trained models (\eg, CLIP~\cite{clip}) to perform few-shot image classification. \cite{clipadapter,tipadapter} further apply a residual feature adapter to improve the few-shot performance.

In this paper, we focus on the zero-shot and few-shot scenarios, where the downstream tasks may not even appear in the pre-training stage. For few-shot learning, we adapt the model with prompt tuning proposed by~\cite{liu2021pre}. The performance of our model can be further improved by fine-tuning the whole model with sufficient downstream training data.

\section{Method}
\label{sec:method}
\begin{figure*}[t]
\centering
\includegraphics[width=0.99\textwidth]{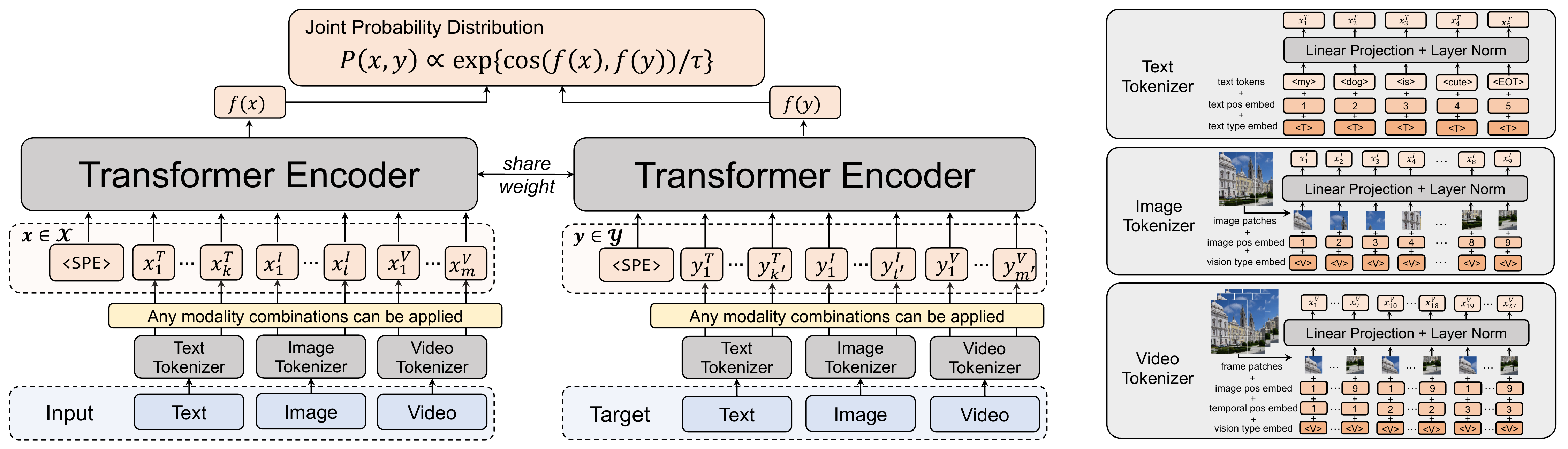}
\caption{Overview of our unified architecture for generic perception. Different task inputs and targets from arbitrary modalities are converted into unified token sequences with modality-specific tokenizers. A modality-agnostic weight-sharing Transformer encoder is then applied to encode these token sequences into the shared representation space. Any perception task can be modeled as finding the maximum likelihood target for each input through the similarity of their representations.}
\label{fig:overview}
\vspace{-0.5em}
\end{figure*}

\subsection{Unified Architecture for Generic Perception}

In this section, we will describe our unified architecture for various modalities and tasks. Fig.~\ref{fig:overview} illustrates the architecture. Specifically, the model first converts different task inputs and targets from arbitrary combinations of modalities into token sequences with modality-specific tokenizers. 
A modality-agnostic Transformer encoder, which shares parameters for different input modalities and target tasks, is then employed to encode different token sequences into a shared representation space.
Any perception task can be modeled in a single unified formulation, which finds the maximum likelihood target for each input through the similarity of their representations.

\vspace{0.5em}\noindent\textbf{Tokenization.~} 
Given the raw inputs from text, image, and video modalities, modality-specific tokenizers are applied to generate the input token sequences for the Transformer encoder.
Here, we use the BPE tokenizer~\cite{sennrich2016neural} for text modality, the image patch tokenizer~\cite{dosovitskiy2020image} for image modality, and the temporal frame patch tokenizer~\cite{bertasius2021space} for video modality. These outputted tokens are attached with additional modality type embeddings to identify which modality the raw input belongs to. Details of the modality-specific tokenizers are described in the Appendix.

As illustrated in Fig.~\ref{fig:overview}, depending on the task requirements, the input sequence $x$ of the Transformer encoder can be composed of different combinations of text token sequence $x^T$, image token sequence $x^I$, and video token sequence $x^V$. 
At the beginning of the sequence $x$, a special token \verb|<SPE>| is always inserted. For example, $x = [\verb|<SPE>|, x^I, x^T]$ for image-text pair inputs, and $x = [\verb|<SPE>|, x^V]$ for video-only inputs, where $[~]$ denotes the sequence concatenation. The feature of \verb|<SPE>| at the encoder output serves as the representation of the input. 

\vspace{0.5em}\noindent\textbf{Generic Modeling of Perception Tasks.~}
We model different perception tasks with a unified architecture, whose parameters are shared for all target tasks.
Each task is defined with a set of inputs $\mathcal{X}$ and a set of candidate targets $\mathcal{Y}$. Given an input $x \in \mathcal{X}$, the task is formulated as finding the maximum likelihood target $y \in \mathcal{Y}$ as
\begin{equation}
\label{eq:inference}
\hat{y} = \arg\max_{y \in \mathcal{Y}} P(x, y),
\end{equation}
where $P(x, y)$ is the joint probability distribution.
The joint probability is estimated through calculating the cosine similarity between the representation of $x$ and $y$ as
\begin{equation}
\label{eq:jointp}
P(x, y) \propto \exp\left(\cos\big(f(x), f(y)\big) / \tau \right),
\end{equation}
where $f(\cdot)$ is the Transformer encoder, and $\tau > 0$ is a learnable temperature parameter.

To obtain generic modeling capability, our unified architecture is pre-trained on a variety of multi-modal tasks simultaneously.  Suppose a series of pre-training tasks is denoted as $\{\mathcal{X}_1, \mathcal{Y}_1\}, \{\mathcal{X}_2, \mathcal{Y}_2\}, ..., \{\mathcal{X}_n, \mathcal{Y}_n\}$, where $\mathcal{X}_i$ and $\mathcal{Y}_i$ is the input set and target set of the $i$-th task, respectively. Then the pre-training loss is defined as
\begin{equation}
L = \sum_{i=1}^{n} \mathop{\mathbb{E}}_{\{x, y\} \in \{\mathcal{X}_i, \mathcal{Y}_i\}} \bigg[- \log \frac{P(x, y)}{\sum_{z \in \mathcal{Y}_i}{P(x, z)}}\bigg],
\end{equation}
where $\mathbb{E}$ is the mathematical expectation, and $\{x, y\} \in \{\mathcal{X}_i, \mathcal{Y}_i\}$ indicates a ground-truth input-target pair sampled from the dataset of the $i$-th task.

Our unified architecture is suitable for any task, as long as its input set $\mathcal{X}$ and target set $\mathcal{Y}$ are composed of images, texts, and videos. For example, the target set $\mathcal{Y}$ in classification tasks can be a set of class names, a set of class descriptions, or even a set of images with handwritten numbers representing class indexes. Detailed instances of $\mathcal{X}$ and $\mathcal{Y}$ will be introduced in the next subsection. Note that we currently focus on text, image, and video modalities, but more modalities are also applicable, as long as the corresponding tokenizers are applied.

\vspace{0.5em}\noindent\textbf{Relation to Previous Perception Models.~}
Our method shares the same goal of learning multi-modal representations as previous perception models. However, existing works follow a task-specific paradigm, while our method is designed for generic perceptual modeling. The main difference lies in two parts: 

1)~ Previous works focus on inputs from certain combinations of modalities required by their target tasks, while our method handles arbitrary combinations of modalities with a unified architecture and shared parameters.

2)~ Previous works require prediction heads specifically designed and trained for each perception task, while our method models different tasks with the same formulation and processes them with unified modeling. 

Therefore, when transferred to a new task, previous methods need to re-design their input formats and prediction heads accordingly. Models require fine-tuning on sufficient task-specific data, resulting in remarkable human and computational costs. In contrast, our method can directly conduct zero-shot inference on novel tasks that do not appear in the pre-training stage. The performance can be further boosted with prompt tuning on few-shot downstream data and fine-tuning on sufficient downstream data.

\begin{figure}[t]
\centering
\includegraphics[width=0.49\textwidth]{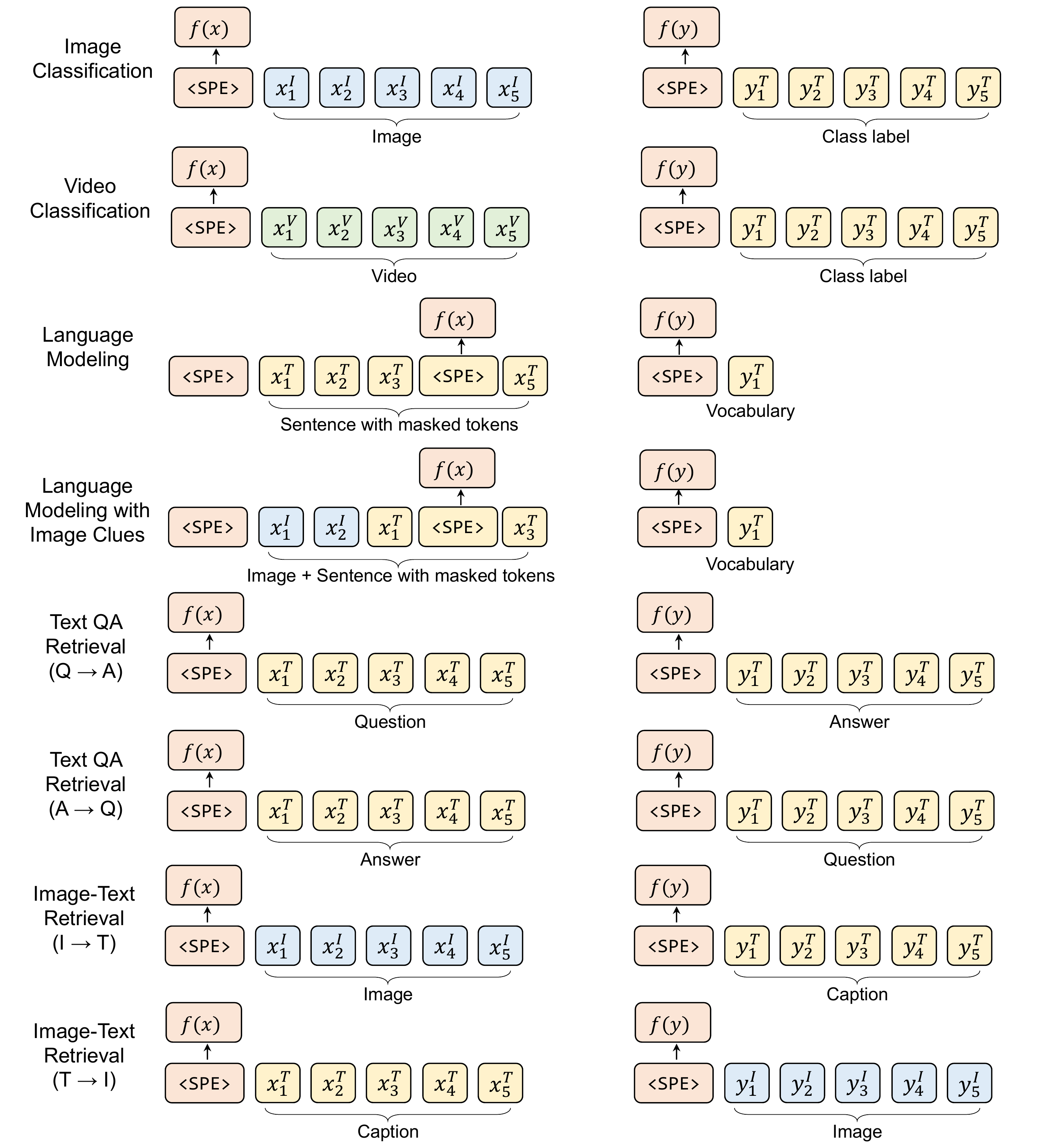}
\caption{Input and target formats of pre-training tasks. For each task, the left column represents the format of input sequence $x$, and the right column represents the format of the target sequence $y$. $f(x)$ and $f(y)$ indicate the representations used for calculating the joint probability distribution as in Eq.~\eqref{eq:jointp}. Here, we have omitted the tokenizer and encoder for concision.}
\label{fig:tasks}
\vspace{-1em}
\end{figure}

\subsection{Pre-training on Multi-Modal Tasks}
Our model is pre-trained on a variety of tasks simultaneously to learn the multi-modal generic representations.
The pre-training tasks are illustrated in Fig.~\ref{fig:tasks}.
Specifically, for uni-modal pre-training tasks, we adopt the most widely-used image classification, video classification, and language modeling tasks.
To further enhance the relationships between different modalities, some cross-modal tasks are also employed, such as language modeling with image clues and image-text retrieval tasks.
Note that for image and video classification tasks, we regard each class name (\eg, \verb|tigershark|) as a text sequence. This provides weak supervision for bridging the gap among the representations of images, videos, and texts. 

\vspace{0.5em}\noindent\textbf{Image and Video Classification.~}
In image and video classification tasks, $\mathcal{X}$ denotes the set of all possible images or videos in the training dataset, and $\mathcal{Y}$ consists of candidate class labels in each dataset. Each class name is regarded as a text sequence to provide weak supervision of the relationship to texts. Both the input $x \in \mathcal{X}$ and target $y \in \mathcal{Y}$ start with an \verb|<SPE>| token, whose feature at the encoder output represents the corresponding sequence.

\vspace{0.5em}\noindent\textbf{Language Modeling with and without Image Clues.~}
The language modeling task aims to predict the masked words according to the context. Both auto-regressive~\cite{Brown2020LanguageMA} and auto-encoding~\cite{devlin2018bert} language modeling are adopted. When inputs have no image, the auto-regressive and auto-encoding tasks correspond to the text generation and the masked language modeling tasks, respectively. 
When inputs have images, the auto-regressive and auto-encoding tasks correspond to the image caption and the masked language modeling with image clues tasks, respectively. 

For auto-encoding language modeling, we follow the practice in BERT~\cite{devlin2018bert} to mask out 15\% words from the text randomly. The model predicts each masked word based on all inputs.
For auto-regressive language modeling, the model predicts each word based on its previous text and image (if any).
Please refer to the Appendix for an efficient implementation of auto-regressive language modeling.

In this task, $\mathcal{X}$ consists of language sentences or image-text pairs. $\mathcal{Y}$ denotes the set of all words in the vocabulary, where each word is regarded as a single text sequence.
Each word that needs to be predicted in $x \in \mathcal{X}$ is replaced by a \verb|<SPE>| token, whose feature at the encoder output is used to match the words in the vocabulary $\mathcal{Y}$. 

\vspace{0.5em}\noindent\textbf{Image and Text Retrieval.~}
For image-text retrieval, the input sets $\mathcal{X}$ and $\mathcal{Y}$ are composed of images and text sequences respectively, or vice versa.
For text-only retrieval, the input sets $\mathcal{X}$ and $\mathcal{Y}$ are both text sequences. Each sequence in $\mathcal{X}$ and $\mathcal{Y}$ has a special token \verb|<SPE>| at the beginning, whose feature at the output of the encoder serves as the final representation.

\subsection{Zero-shot, Prompt Tuning and Fine-tuning}
During the pre-training stage, our unified architecture learns to model the joint distribution of input and target sequences from arbitrary modalities.
Thanks to the generic perceptual modeling, our pre-trained model can perform zero-shot inference on completely novel tasks that do not appear in the pre-training stage.
Our model can be further adapted to downstream tasks with task-specific additional training data. For the few-shot scenario, we employ the prompt tuning~\cite{liu2021pre} scheme, which only adds a few additional task-specific parameters to the model. The performance on specific tasks can be further improved by fine-tuning the whole model on sufficient downstream data.

\vspace{0.5em}\noindent\textbf{Zero-shot Inference on Novel Tasks.~}
Our model has the potential to perform zero-shot inference on any perception task that can be modeled by a joint probability distribution. For a task with input $x \in \mathcal{X}$ and a candidate target $y \in \mathcal{Y}$, we firstly tokenize $x$ and $y$ into two sequences. The joint probability $P(x,y)$ is then estimated following Eq.~\eqref{eq:jointp}. Zero-shot inference can be conducted by maximum likelihood estimation, as described in Eq.~\eqref{eq:inference}. 
Performance can also be improved through prompt engineering, similar to the prompting~\cite{liu2021pre} for language models such as GPT-3~\cite{Brown2020LanguageMA}, where network training is not required.

\vspace{0.5em}\noindent\textbf{Prompt Tuning.~}
For the few-shot scenario with limited training data, we adopt prompt tuning, which is memory-efficient and has been proved to be better than the fine-tuning scheme in few-shot NLP~\cite{gao2021making}.
In prompt tuning, most pre-trained parameters are fixed, leaving only a small portion of task-specific parameters to be optimized.
Specifically, following P-Tuning v2~\cite{liuPtuningv2}, learnable prompt tokens with random initialization are added at each layer of the Transformer encoder, and class labels with linear heads are added for classification tasks. The \verb|<SPE>| token and layer norm parameters are also tuned. We refer the readers to the Appendix for more details.

\vspace{0.5em}\noindent\textbf{Fine-Tuning.~} 
For downstream tasks with sufficient training data, our model can also be fine-tuned to further improve its performance. During fine-tuning, our model can serve as a joint probability estimator (same as our proposed generic perceptual modeling), or a feature extractor (same as traditional pre-trained models).
Under the setting of joint probability estimation, the downstream tasks are formulated in the same unified manner as in pre-training. On the other hand, similar to previous perception models, our model can also be used as a feature extractor by adding a task-specific head on the top of the encoder.
We empirically find these two schemes achieve very similar performance, and hence the scheme of joint probability distribution estimator is used by default for consistency.

\section{Experiments}
\label{sec:exp}
\subsection{Datasets}

\begin{table}[t]
\small
    \centering
\resizebox{0.37\textwidth}{!}{
\begin{tabular}{c|ccc}

\toprule
Dataset  & $\#$Images & $\#$Videos& $\#$Text \\

\midrule

ImageNet-21k~\cite{deng2009imagenet}  & 14.2M & 0 & 21K\\
Kinetics-700~\cite{kay2017kinetics} & 0 &  542K&  700\\
Moments in Time~\cite{monfort2019moments} & 0 & 792K & 339 \\
Books\&Wiki~\cite{zhu2015aligning}  & 0 & 0 &  101M\\
PAQ~\cite{lewis2021paq} & 0 & 0 & 65M  \\
CC3M~\cite{sharma2018conceptual}  & 3.0M & 0 & 3.0M  \\
CC12M~\cite{changpinyo2021cc12m}  & 11.1M & 0 & 11.1M  \\
COCO Caption~\cite{Chen2015MicrosoftCC}  & 113K & 0  & 567K  \\
Visual Genome~\cite{krishna2017visual}   & 108K  & 0 & 5.41M  \\
SBU~\cite{ordonez2011im2text} & 830K & 0 & 830K  \\
YFCC\footnote{We use the filtered version of YFCC100M proposed by CLIP~\cite{clip}.}~\cite{yfcc} & 14.8M & 0 & 14.8M  \\
\bottomrule
\end{tabular}}
    \caption{Pre-training dataset statistics. $\#$Images, $\#$Videos and $\#$Text represent the number of images, video clips, and textual sentences (or phrases), respectively.
    } 
    \label{tab:dataset_statistics}
    \vspace{-1.2em}
\end{table}
Our model is pre-trained on a variety of tasks, whose statistics are listed in Tab.~\ref{tab:dataset_statistics}. We pre-train image classification on ImageNet-21k~\cite{deng2009imagenet}. For video classification, we pre-train on Kinetics-700~\cite{kay2017kinetics} and Moments in Time~\cite{monfort2019moments}. We pre-train language modeling on BookCorpora~\cite{zhu2015aligning} \& English Wikipedia (Books\&Wiki) and PAQ~\cite{lewis2021paq}. For language modeling with image clues and image-text retrieval, we use a combination of COCO Caption~\cite{chen2015microsoft}, SBU Captions (SBU)~\cite{ordonez2011im2text}, Visual Genome~\cite{krishna2017visual}, CC3M~\cite{sharma2018conceptual}, CC12M~\cite{changpinyo2021cc12m} and YFCC~\cite{yfcc}. 
To evaluate the effectiveness of our method and verify the generalization of our pre-trained model, we also use several novel datasets that did not appear in pre-training, \ie, Flickr30k \cite{plummer2015flickr30k}, MSVD \cite{msvd}, VQA \cite{balanced_vqa_v2}, and GLUE \cite{wang2018glue}.
See Appendix for the details of datasets.

\subsection{Implementation Details}
The Transformer encoder used for experiments is of the same configuration with BERT\textsubscript{BASE}~\cite{devlin2018bert}. It is a 12-layer encoder with the embedding dimension of 768 and the attention head number of 12. The hidden dimension size in FFN is 3072. 
We pre-train the model with multiple tasks simultaneously. In each iteration, each GPU independently samples a single task and dataset. The gradients of different GPUs are synchronized after the gradient back-propagation. 
We use AdamW~\cite{kingma2014adam} optimizer with a base learning rate of $0.0002$ and a weight decay of $0.05$. Gradient clipping with 5.0 is used to stabilize training.
We also use drop path~\cite{Larsson2017FractalNetUN} with a probability of 0.1 during training. 
The model is pre-trained on 128 Tesla V100 GPUs in a distributed fashion for 500k iterations.  We use the cosine learning rate schedule with 50k iterations of linear warmup.
See Appendix for more implementation details.
 
\begin{table} [t]
    \centering
\resizebox{0.38\textwidth}{!}{

    \begin{tabular}{P{7em} P{6em} P{7em} }
    \toprule
 \multirow{2}{*}{Task}     & ImageNet-1k & Kinetics-400 \\
  & Acc & Acc \\
  \midrule
      DeiT~\cite{touvron2021training}   & 81.8 & - \\
      TimeSformer~\cite{bertasius2021space} & - & 75.5 \\ 
  \midrule
       Ours \textsubscript{w/o Tuning} & 78.0 & 73.5 \\
  \midrule
       Ours \textsubscript{PT (0.1\%)} & 79.4 & 73.6 \\
       Ours \textsubscript{FT (0.1\%)} & 78.8 & 73.5 \\
  \midrule
       Ours \textsubscript{PT (1\%)} & 80.2 & 73.6 \\
       Ours \textsubscript{FT (1\%)} & 80.2 & 73.6 \\
  \midrule
       Ours \textsubscript{FT (100\%)} & 83.8 & 75.8 \\
   \bottomrule   
    \end{tabular}
}
    \caption{Image and video classification performance under different tuning settings. PT means prompt-tuning, and FT means fine-tuning. The percentage of data used in tuning is noted.}
    \label{tab:classification}
    \vspace{-1.2em}
\end{table}
\begin{table*} [t]
    \centering
\resizebox{0.79\textwidth}{!}{
    \begin{tabular}{c|cccccc|cccccc}
    \toprule
 \multirow{3}{*}{Task}  & \multicolumn{6}{c|}{Text Retrieval} & \multicolumn{6}{c}{ Image Retrieval} \\
  &  \multicolumn{3}{c}{Flickr30k} &   \multicolumn{3}{c|}{COCO Caption} &  \multicolumn{3}{c}{Flickr30k} &   \multicolumn{3}{c}{COCO Caption}  \\
   &   R@1 & R@5 & R10 & R@1 & R@5 & R10 & R@1 & R@5 & R10 & R@1 & R@5 & R10 \\
   \midrule
   ImageBERT~\cite{Qi2020ImageBERTCP}  \textsubscript{w/o Tuning} & 70.7 & 90.2 & 94.0 & 44.0 & 71.2 & 80.4 & 54.3 & 79.6 & 87.5 & 32.3 & 59.0 & 70.2 \\
   UNITER-B~\cite{chen2020uniter} \textsubscript{w/o Tuning} & 80.7 & 95.7 & 98.0 &  -  & - &- & 66.2 & 88.4 & 92.9 & - & -& -\\
   ViLT~\cite{kim2021vilt} \textsubscript{w/o Tuning} & 73.2 & 93.6 & 96.5 & 56.5 & 82.6 & 89.6 & 55.0 & 82.5 & 89.8 & 40.4 & 70.0 & 81.1  \\
   
   \midrule
   Unicoder-VL~\cite{huang2019unicoder} & 86.2 & 96.3 & 99.0 &  62.3 & 87.1 & 92.8 & 71.5 & 91.2 & 95.2 &  48.4 & 76.7 & 85.9 \\
   UNITER-B & 85.9 & 97.1 & 98.8 & 64.4 & 87.4 & 93.1 & 72.5 & 92.4 & 96.1 & 50.3 & 78.5 & 87.2  \\
   ViLT & 83.5 & 96.7 & 98.6 & 61.5 & 86.3 & 92.7 & 64.4 & 88.7 & 93.8 & 42.7 & 72.9 & 83.1  \\ 
   
   \midrule
     
    Ours \textsubscript{w/o Tuning}  & 74.8 & 94.8 & 98.2 & 57.7 & 85.6& 92.3&65.8 & 88.8& 93.6 & 46.3 & 75.0 & 84.0 \\   
    \midrule
    Ours \textsubscript{PT (1\%)}  & 84.4   & 97.8& 99.2 & 61.4 & 86.7 & 93.2 & 71.1 & 91.6 & 95.1 &  47.0 & 75.3 & 84.3    \\
    Ours \textsubscript{FT (1\%)}  & 78.4 & 95.7 & 97.8 & 60.2      &85.1 & 90.6   &61.0 & 85.7 & 91.0 & 43.6 & 70.9 & 80.5 \\
    
    \midrule
    Ours \textsubscript{PT (10\%)} & 86.4 & 98.2 & 99.5 & 61.6 & 87.0 & 93.2 & 72.5 & 92.3 & 95.7 & 47.2 & 75.4 & 84.3      \\
    Ours \textsubscript{FT (10\%)}  & 84.9 & 97.4 & 98.3 & 60.9& 85.5 & 92.1 &  67.9 & 89.4 & 92.9 &  45.6 & 73.4 & 82.6                  \\
    \midrule
     Ours \textsubscript{FT (100\%)} & 87.9 & 98.2 & 99.1 & 64.7 & 87.8 & 93.7 & 74.9 & 93.5 & 96.0 & 48.3 & 75.9 & 84.5  \\
 
   \bottomrule   
    \end{tabular}

}
    \vspace{-0.25em}
    \caption{Image-text retrieval performance under different tuning  settings. PT means prompt-tuning, and FT means fine-tuning. The percentage of data used in tuning is noted.}
    \label{tab:img_retrieval}
    \vspace{-0.5em}
\end{table*}

\begin{table} [t]
    \centering
\resizebox{0.48\textwidth}{!}{

    \begin{tabular}{c|cccc|cccc }
    \toprule
 \multirow{2}{*}{Task}     & \multicolumn{4}{c|}{COCO Caption} & \multicolumn{4}{c}{ Flickr30k} \\
 
    & B@4 & M & C & S & B@4 & M & C & S  \\
   \midrule
   Unified VLP \cite{zhou2020unified}  & 36.5 & 28.4 & 116.9 & 21.2 & 30.1 & 23.0 & 67.4 & 17.0   \\

   \midrule
     
   Ours \textsubscript{w/o Tuning} &  33.6 & 27.0 &  109.8 & 20.3 & 17.0 & 16.2 & 41.2 & 11.2  \\
   \midrule
   Ours \textsubscript{PT (1\%)}  & 34.3 & 27.2 &109.6& 21.2 & 28.1& 21.6 & 59.1 & 15.6  \\
  Ours \textsubscript{FT (1\%)}  & 28.0 & 26.8 & 100.1 & 20.2 & 18.9 & 19.7 & 45.3 & 14.3\\
   
   \midrule

     Ours \textsubscript{PT (10\%)} & 35.0 & 27.9 & 114.1& 21.3 &28.8 & 22.1 & 61.7 & 16.8  \\
   Ours \textsubscript{FT (10\%)} & 32.7 & 27.5 & 109.0 &  21.1 & 26.9 & 21.6 & 52.1 & 14.5 \\
   \midrule
   Ours \textsubscript{FT (100\%)} & 35.6  & 28.1 & 116.5 & 21.5 & 30.1 & 24.5 & 72.7 & 18.2\\
 
   \bottomrule   
    \end{tabular}
}
    \vspace{-0.25em}
    \caption{Image caption performance under different tuning settings. B@4, M, C, S stand for BLEU-4, METEOR, CIDEr, and SPICE scores, respectively.}
    \label{tab:img_caption}
    \vspace{-0.25em}
\end{table}

\subsection{Evaluation on Pre-training Tasks}
We first evaluate our pre-trained model on tasks that have been involved in the pre-training stage, while the datasets might be different.
The widely used Imagenet-1k~\cite{deng2009imagenet} and Kinetics-400~\cite{kay2017kinetics} are used for evaluating the image and video classification tasks, respectively.
COCO Caption and Flickr30k are the typical datasets used to evaluate the performance on image caption and image-text retrieval. 

\vspace{0.5em}\noindent\textbf{Results.~}
Tab.~\ref{tab:classification}, Tab.~\ref{tab:img_retrieval}, and Tab.~\ref{tab:img_caption} present the evaluation results of our models on four pre-training tasks, \ie, image classification, video classification, image-text retrieval, and image caption. We compare our model with task-specific SOTA methods having the similar model size. 

Results show that without any tuning, our pre-trained model reaches reasonable performance on these tasks. Although the performance is slightly worse than the SOTA methods. We speculate that the performance gap is due to the limited capacity of our model, which may have a negative impact on the representation ability. Note that our method shares a similar model size with other methods, but need to simultaneously process much more pre-training tasks from various datasets and modalities.

By conducting prompt tuning on each task with 1\% downstream data, the performance is boosted to a level close to SOTA performance. It's worth noting that all parameters of other methods are specifically trained on the target tasks. While for our prompt tuning, only a small amount of parameters are tuned, and the encoder is still fixed and shared among different tasks, indicating that our method can handle different tasks with low marginal cost. 

We further fine-tune the pre-trained model with 100\% of the downstream data. With full-data fine-tuning, our model achieves performance on-par with or better than the SOTA methods on all these tasks, which proves our model has learned high-quality representations. 
We also compare the performance of prompt tuning and fine-tuning in the scenario of few-shot learning. On all of these tasks, prompt tuning shows a consistently better performance than fine-tuning with the same amount of data, which demonstrates its superiority under few-shot scenarios.

\begin{table} [t]
    \centering

\resizebox{0.38\textwidth}{!}{
    \begin{tabular}{P{7em} P{2em} P{2em}  P{2em} P{2em} P{2em}}
    \toprule
  \multirow{2}{*}{Task}  & \multicolumn{5}{c}{MSVD} \\
  &  B@4 & M & R&  C & S \\
  \midrule
      ORG-TRL \cite{zhang2020object} &   54.3 & 36.4 & 73.9 & 95.2 & - \\
      \midrule 
      Ours \textsubscript{w/o Tuning} & 20.3 & 25.8 & 52.1 & 45.7 & 6.5 \\ 
      \midrule
      Ours \textsubscript{PT (1\%)} & 54.8 & 38.9 & 74.7 &  104.8 & 6.6 \\ 
       Ours \textsubscript{FT (1\%)} &  47.3 & 35.8 & 66.2 &80.1 & 6.2\\
      \midrule

       Ours  \textsubscript{PT (10\%)}  &  57.2 & 39.1 & 75.6 & 112.1 & 6.8 \\
      Ours  \textsubscript{FT (10\%)}  & 56.7 &38.7 &  70.0 & 88.2 & 6.7 \\
      \midrule
      Ours  \textsubscript{FT (100\%)}  & 61.5 & 42.3 & 79.0 & 131.0 & 7.7 \\ 
   \bottomrule   
    \end{tabular}
}

    \caption{Video caption (novel task) performance under different tuning settings. Note that this task did not appear in our pre-training. The only task related to video modality in our pre-training is video classification.}
    \label{tab:msvd_caption}
    \vspace{-1.3em}
\end{table}

\begin{table} [t]
    \centering
    \small 
\resizebox{0.45\textwidth}{!}{
    \begin{tabular}{c|ccc|ccc}
    \toprule
\multirow{3}{*}{Task}    & \multicolumn{3}{c|}{Text Retrieval} & \multicolumn{3}{c}{ Video Retrieval}\\
& \multicolumn{3}{c|}{MSVD} & \multicolumn{3}{c}{MSVD}\\
  & R@1 & R@5 & R10 & R@1  & R@5 & R@10   \\
  \midrule
     
      CLIP4clip \cite{luo2021clip4clip} & 56.6 & 79.7 & 84.3 &46.2 & 76.1 & 84.6 \\
      CLIP2video \cite{fang2021clip2video} & 58.7 & 85.6 & 91.6 & 47.0 & 76.8 & 85.9 \\

      \midrule 
      Ours \textsubscript{w/o Tuning} & 42.7 & 69.1 & 79.6 & 34.6 & 64.5 & 75.4 \\ 
      \midrule
     Ours  \textsubscript{PT (1\%)} & 61.2 & 83.7 & 89.0 & 42.6 & 73.3 & 82.5 \\ 
      Ours  \textsubscript{FT (1\%)} & 49.6 & 75.8 & 83.7 & 37.5 & 68.2 & 79.3  \\ 
      
      \midrule
      
      Ours \textsubscript{PT (10\%)}) & 61.3 & 84.8 & 90.9 & 43.1 & 74.2 & 83.4 \\ 
      Ours \textsubscript{FT (10\%)}) & 59.1 & 81.9 & 87.4 & 41.7 &  71.6 & 81.3\\ 
      \midrule
      Ours \textsubscript{FT (100\%)} & 61.5 & 83.5 & 90.2 & 45.4 & 75.8 & 85.0 \\ 
   \bottomrule   
    \end{tabular}
}
    \caption{Video-text retrieval (novel task) performance under different tuning settings. Note that this task did not appear in our pre-training. }
    \label{tab:msvd_retrieval}
    \vspace{-0.5em}
\end{table}

\subsection{Generalization to Novel Tasks}
Thanks to the generic perceptual modeling, our pre-trained model can generalize to novel tasks by converting the tasks into our unified task formulation. We evaluate zero-shot inference on tasks that did not appear in pre-training, \ie, video caption, video-text retrieval, visual question answering, and natural language understanding.

\begin{table} [t]
    \centering
    \small 
\resizebox{0.33\textwidth}{!}{
    \begin{tabular}{c|ccc}
    \toprule
 \multirow{2}{*}{Task}      & \multicolumn{3}{c}{VQA v2 test-dev}  \\
   & Yes/No & Numbers & Others \\
  \midrule
      Unified VLP \cite{zhou2020unified} & 87.2 & 52.1 & 60.3 \\
      \midrule 
      Ours \textsubscript{w/o Tuning}& 0.9 & 3.0 & 25.5 \\ 
      \midrule
      Ours \textsubscript{PT (0.1\%)} & 63.0 & 31.8 & 49.6 \\ 
     Ours \textsubscript{FT (0.1\%)} & 63.0 & 30.6 & 49.1\\ 
      \midrule 
      
       Ours \textsubscript{PT (1\%)} & 70.8 & 41.3 & 57.7\\ 
      Ours \textsubscript{FT (1\%)} & 71.0 & 42.4 & 57.5\\ 
      \midrule
      Ours \textsubscript{FT (100\%)} & 84.8 & 47.4 & 61.8\\ 
   \bottomrule   
    \end{tabular}
}
    \caption{Visual question answering (novel task) performance under different tuning settings. Note that this task did not appear in our pre-training. }
    \label{tab:vqa}
    \vspace{-1.5em}
\end{table}

\vspace{0.4em}\noindent\textbf{Video Caption and Video-Text Retrieval.~}
Our pre-trained model is evaluated on MSVD~\cite{msvd} dataset.
Specifically, for video caption, $ \mathcal{X}_1 $ consists of the concatenation of video and language sequences that have been predicted, and  $ \mathcal{X}_2$ denotes the set of all words in the vocabulary.
For the video-text retrieval, the input sets $\mathcal{X}_1$ and $\mathcal{X}_2$ consist of possible video and text sequences, or vice versa. 
\begin{table} [t]
    \centering
    \small 
\resizebox{0.45\textwidth}{!}{
    \begin{tabular}{c|cccccc}
    \toprule

\multirow{3}{*}{Task}  & \multicolumn{6}{c}{GLUE} \\
 & MNLI & QNLI & QQP     & RTE &  SST-2 &  MRPC  \\
   & (Acc) & (Acc) & (F1) & (Acc) & (Acc) & (F1) \\
  \midrule
 
PLM  \cite{anonymous2022are}~\textsubscript{w/o tuning}   & 49.4 & 50.7 & 46.6 & 53.8 & 70.6 & 44.2 \\
      BERT\textsubscript{BASE} \cite{wang2018glue}  & 84.6 & 92.7 & 71.2 & 66.4 & 93.5 & 88.9 \\

      \midrule 
      Ours \textsubscript{w/o Tuning} & 49.6 & 51.0 & 53.6 & 55.6 & 70.6 & 76.1  \\ 
      \midrule
      Ours \textsubscript{PT (1\%)}  & 60.1 & 76.0 &  70.2 & 56.3 & 80.9 & 80.3 \\ 
       Ours \textsubscript{FT (1\%)}  & 47.3 & 60.6 & 68.9 & 49.1 & 69.7 & 72.3 \\ 
      
     \midrule
      Ours \textsubscript{PT (10\%)}  & 68.5 & 83.2 & 77.0 & 58.2 & 83.4 & 83.2 \\ 
     
      Ours \textsubscript{FT (10\%)}  & 60.5 & 71.5 & 71.4 & 50.5 &  79.1 & 80.6  \\  \midrule
     
      Ours \textsubscript{FT (100\%)}  & 81.7 & 89.9 & 87.1 & 64.3 & 90.2 & 86.6  \\ 
   \bottomrule   
    \end{tabular}
}
    \caption{Natural language understanding (novel task) performance under different tuning settings. Note that this task did not appear in our pre-training. }
    \label{tab:glue}
    \vspace{-1.4em}
\end{table}

\vspace{0.4em}\noindent\textbf{Visual Question Answering.~} 
In visual question answering, the model is asked to answer a question w.r.t a reference image from a list of answer candidates. 
We evaluate our pre-trained model on VQA~\cite{balanced_vqa_v2} dataset. $\mathcal{X}_1$ is a set of image-text sequence, where the text is the question tokens followed by a \verb|<SPE>| token used to predict the answers. Each $x_2 \in \mathcal{X}_2$ is an answer sequence beginning with \verb|<SPE>|. Inference is achieved by computing the similarity between output features of \verb|<SPE>| in $x_1$ and $x_2$.

\vspace{0.4em}\noindent\textbf{Natural Language Understanding.~}
Six language-only tasks are chosen from GLUE benchmark~\cite{wang2018glue} to evaluate the natural language understanding ability of our pre-trained model.
These tasks are either single sentence classification or sentence-pair classification tasks. We follow \cite{gao2020making} to construct the textual class labels for each dataset. Here, the input sequence $x_1 \in \mathcal{X}_1 $ denotes the input single sentence or the sentence-pair,  and the sequence $ x_2 \in \mathcal{X}_2 $ represents the  class labels in each dataset.

\vspace{0.5em}\noindent\textbf{Result.~}
Tab.~\ref{tab:msvd_caption}, Tab.~\ref{tab:msvd_retrieval} and Tab.~\ref{tab:vqa} show the results on video caption and video-text retrieval and visual question answering, respectively. Our pre-trained model can obtain reasonable zero-shot performance on these novel tasks. Note that none of previous works can perform this type of zero-shot inference at all. From Tab.~\ref{tab:vqa}, we note that our model shows unsatisfactory zero-shot performance on ``Yes/No" and ``Number" subsets in VQA. We speculate that it may be due to the distribution difference between those answers and our pre-training corpora. We futher conduct prompt tuning on these tasks with only $ 1\% $ data, which brings our model to a level close to the SOTA results. By further fine-tuning with 100\% downstream data, our model can achieve results on par with or better than the SOTA methods.

On the GLUE benchmark, our model can achieve comparable performance with \cite{anonymous2022are} in zero-shot evaluation.
When fine-tuning the pre-trained model with 100\% downstream data, our model performs slightly worse than BERT\textsubscript{BASE}.
Since our model has the same number of parameters as BERT\textsubscript{BASE}, but need to process much more tasks from various datasets and modalities, we speculate that the performance drop is due to the limited capacity of the model.

\section{Conclusion}
\label{sec:conclusion}
In this paper, we propose a unified perception architecture that processes various modalities and tasks with a single model and shared parameters. With pre-training on uni-modal and multi-modal tasks, our model shows the ability of zero-shot inference on novel tasks, and can reach the performance close to SOTA results by prompt tuning with only a small amount of downstream data. The performance can be further 
improved to be on par with or superior to SOTA results by full-data fine-tuning.

\vspace{0.4em}
\noindent\textbf{Limitations.} Our method is currently only applicable when the target set is discrete, such as classification and retrieval. Whether our model can be extended to regression tasks is still questionable. Future work may explore the unified perception model of both discrete and continuous target sets.

\vspace{0.4em}
\noindent\textbf{Potential Negative Societal Impact.} This work may share the common negative impacts of large-scale training, which may consume lots of electricity and result in increased carbon emissions. This method also learns from a large number of datasets that may contain data biases. Future work may seek for more efficient and unbiased training.

\paragraph{Acknowledgments} The work is supported by the National Key R\&D Program of China (2020AAA0105200), Beijing Academy of Artificial Intelligence.

{\small
\bibliographystyle{ieee_fullname}
\bibliography{egbib}

\begin{thebibliography}{10}\itemsep=-1pt

\bibitem{cubuk2020randaugment}
Ekin~D Cubuk, Barret Zoph, Jonathon Shlens, and Quoc~V Le.
\newblock Randaugment: Practical automated data augmentation with a reduced
  search space.
\newblock In {\em CVPR Workshops}, pages 702--703, 2020.

\bibitem{mrpc}
William~B Dolan and Chris Brockett.
\newblock Automatically constructing a corpus of sentential paraphrases.
\newblock In {\em Proceedings of the Third International Workshop on
  Paraphrasing (IWP2005)}, 2005.

\bibitem{flickr2020terms}
Inc. Flickr.
\newblock Flickr terms \& conditions of use.
\newblock \url{https://www.flickr.com/help/terms}.

\bibitem{vgterms}
Ranjay Krishna.
\newblock Visual genome terms \& conditions of use.
\newblock \url{https://visualgenome.org/about}.

\bibitem{cc12mlicense}
Google LLC.
\newblock Conceptual 12m terms \& conditions of use.
\newblock
  \url{https://github.com/google-research-datasets/conceptual-12m/blob/main/LICENSE}.

\bibitem{cc3mlicense}
Google LLC.
\newblock Conceptual captions terms \& conditions of use.
\newblock
  \url{https://github.com/google-research-datasets/conceptual-captions/blob/master/LICENSE}.

\bibitem{Smaira2020ASN}
Lucas Smaira, Jo{\~a}o Carreira, Eric Noland, Ellen Clancy, Amy Wu, and Andrew
  Zisserman.
\newblock A short note on the kinetics-700-2020 human action dataset.
\newblock {\em arXiv preprint arXiv:2010.10864}, 2020.

\bibitem{SST2}
Richard Socher, Alex Perelygin, Jean Wu, Jason Chuang, Christopher~D Manning,
  Andrew~Y Ng, and Christopher Potts.
\newblock Recursive deep models for semantic compositionality over a sentiment
  treebank.
\newblock In {\em Proceedings of the 2013 conference on empirical methods in
  natural language processing}, pages 1631--1642, 2013.

\bibitem{imagenetterms}
Princeton University and Stanford University.
\newblock Imagenet terms \& conditions of use.
\newblock \url{https://image-net.org/download}.

\bibitem{yun2019cutmix}
Sangdoo Yun, Dongyoon Han, Seong~Joon Oh, Sanghyuk Chun, Junsuk Choe, and
  Youngjoon Yoo.
\newblock Cutmix: Regularization strategy to train strong classifiers with
  localizable features.
\newblock In {\em ICCV}, pages 6023--6032, 2019.

\bibitem{zhang2017mixup}
Hongyi Zhang, Moustapha Cisse, Yann~N Dauphin, and David Lopez-Paz.
\newblock mixup: Beyond empirical risk minimization.
\newblock {\em arXiv preprint arXiv:1710.09412}, 2017.

\bibitem{zhong2020random}
Zhun Zhong, Liang Zheng, Guoliang Kang, Shaozi Li, and Yi Yang.
\newblock Random erasing data augmentation.
\newblock In {\em AAAI}, volume~34, pages 13001--13008, 2020.

\end{thebibliography}


\begin{thebibliography}{10}\itemsep=-1pt

\bibitem{Akbari2021VATTTF}
Hassan Akbari, Li Yuan, Rui Qian, Wei-Hong Chuang, Shih-Fu Chang, Yin Cui, and
  Boqing Gong.
\newblock Vatt: Transformers for multimodal self-supervised learning from raw
  video, audio and text.
\newblock {\em arXiv preprint arXiv:2104.11178}, 2021.

\bibitem{b2t2}
Chris Alberti, Jeffrey Ling, Michael Collins, and David Reitter.
\newblock Fusion of detected objects in text for visual question answering.
\newblock {\em arXiv preprint arXiv:1908.05054}, 2019.

\bibitem{anonymous2022are}
Anonymous.
\newblock Are {BERT} families zero-shot learners? a study on their potential
  and limitations.
\newblock In {\em Submitted to ICLR}, 2022.
\newblock under review.

\bibitem{arnab2021vivit}
Anurag Arnab, Mostafa Dehghani, Georg Heigold, Chen Sun, Mario Lu{\v{c}}i{\'c},
  and Cordelia Schmid.
\newblock Vivit: A video vision transformer.
\newblock {\em arXiv preprint arXiv:2103.15691}, 2021.

\bibitem{gberta_2021_ICML}
Gedas Bertasius, Heng Wang, and Lorenzo Torresani.
\newblock Is space-time attention all you need for video understanding?
\newblock In {\em ICML}, July 2021.

\bibitem{bertasius2021space}
Gedas Bertasius, Heng Wang, and Lorenzo Torresani.
\newblock Is space-time attention all you need for video understanding?
\newblock {\em arXiv preprint arXiv:2102.05095}, 2021.

\bibitem{Brown2020LanguageMA}
Tom~B. Brown, Benjamin Mann, Nick Ryder, Melanie Subbiah, Jared Kaplan,
  Prafulla Dhariwal, Arvind Neelakantan, Pranav Shyam, Girish Sastry, Amanda
  Askell, Sandhini Agarwal, Ariel Herbert-Voss, Gretchen Krueger, T.~J.
  Henighan, Rewon Child, Aditya Ramesh, Daniel~M. Ziegler, Jeff Wu, Clemens
  Winter, Christopher Hesse, Mark Chen, Eric Sigler, Mateusz Litwin, Scott
  Gray, Benjamin Chess, Jack Clark, Christopher Berner, Sam McCandlish, Alec
  Radford, Ilya Sutskever, and Dario Amodei.
\newblock Language models are few-shot learners.
\newblock {\em arXiv preprint arXiv:2005.14165}, 2020.

\bibitem{changpinyo2021cc12m}
Soravit Changpinyo, Piyush Sharma, Nan Ding, and Radu Soricut.
\newblock {Conceptual 12M}: Pushing web-scale image-text pre-training to
  recognize long-tail visual concepts.
\newblock In {\em CVPR}, 2021.

\bibitem{chen2021crossvit}
Chun-Fu Chen, Quanfu Fan, and Rameswar Panda.
\newblock Crossvit: Cross-attention multi-scale vision transformer for image
  classification.
\newblock {\em arXiv preprint arXiv:2103.14899}, 2021.

\bibitem{msvd}
David Chen and William~B Dolan.
\newblock Collecting highly parallel data for paraphrase evaluation.
\newblock In {\em ACL}, pages 190--200, 2011.

\bibitem{Chen2015MicrosoftCC}
Xinlei Chen, Hao Fang, Tsung-Yi Lin, Ramakrishna Vedantam, Saurabh Gupta, Piotr
  Doll{\'a}r, and C.~Lawrence Zitnick.
\newblock Microsoft coco captions: Data collection and evaluation server.
\newblock {\em arXiv preprint arXiv:1504.00325}, 2015.

\bibitem{chen2015microsoft}
Xinlei Chen, Hao Fang, Tsung-Yi Lin, Ramakrishna Vedantam, Saurabh Gupta, Piotr
  Doll{\'a}r, and C~Lawrence Zitnick.
\newblock Microsoft coco captions: Data collection and evaluation server.
\newblock {\em arXiv preprint arXiv:1504.00325}, 2015.

\bibitem{chen2019catastrophic}
Xinyang Chen, Sinan Wang, Bo Fu, Mingsheng Long, and Jianmin Wang.
\newblock Catastrophic forgetting meets negative transfer: Batch spectral
  shrinkage for safe transfer learning.
\newblock In {\em NeurIPS}, 2019.

\bibitem{chen2020uniter}
Yen-Chun Chen, Linjie Li, Licheng Yu, Ahmed El~Kholy, Faisal Ahmed, Zhe Gan, Yu
  Cheng, and Jingjing Liu.
\newblock Uniter: Universal image-text representation learning.
\newblock In {\em ECCV}, pages 104--120. Springer, 2020.

\bibitem{deng2009imagenet}
Jia Deng, Wei Dong, Richard Socher, Li-Jia Li, Kai Li, and Li Fei-Fei.
\newblock Imagenet: A large-scale hierarchical image database.
\newblock In {\em CVPR}, pages 248--255. Ieee, 2009.

\bibitem{devlin2018bert}
Jacob Devlin, Ming-Wei Chang, Kenton Lee, and Kristina Toutanova.
\newblock Bert: Pre-training of deep bidirectional transformers for language
  understanding.
\newblock {\em arXiv preprint arXiv:1810.04805}, 2018.

\bibitem{dosovitskiy2020image}
Alexey Dosovitskiy, Lucas Beyer, Alexander Kolesnikov, Dirk Weissenborn,
  Xiaohua Zhai, Thomas Unterthiner, Mostafa Dehghani, Matthias Minderer, Georg
  Heigold, Sylvain Gelly, et~al.
\newblock An image is worth 16x16 words: Transformers for image recognition at
  scale.
\newblock {\em arXiv preprint arXiv:2010.11929}, 2020.

\bibitem{fang2021clip2video}
Han Fang, Pengfei Xiong, Luhui Xu, and Yu Chen.
\newblock Clip2video: Mastering video-text retrieval via image clip.
\newblock {\em arXiv preprint arXiv:2106.11097}, 2021.

\bibitem{gao2020making}
Tianyu Gao, Adam Fisch, and Danqi Chen.
\newblock Making pre-trained language models better few-shot learners.
\newblock {\em arXiv preprint arXiv:2012.15723}, 2020.

\bibitem{gao2021making}
Tianyu Gao, Adam Fisch, and Danqi Chen.
\newblock Making pre-trained language models better few-shot learners.
\newblock In {\em ACL}, 2021.

\bibitem{balanced_vqa_v2}
Yash Goyal, Tejas Khot, Douglas Summers{-}Stay, Dhruv Batra, and Devi Parikh.
\newblock Making the {V} in {VQA} matter: Elevating the role of image
  understanding in {V}isual {Q}uestion {A}nswering.
\newblock In {\em CVPR}, 2017.

\bibitem{hu2021unit}
Ronghang Hu and Amanpreet Singh.
\newblock Unit: Multimodal multitask learning with a unified transformer.
\newblock {\em arXiv preprint arXiv:2102.10772}, 2021.

\bibitem{huang2019unicoder}
Haoyang Huang, Yaobo Liang, Nan Duan, Ming Gong, Linjun Shou, Daxin Jiang, and
  Ming Zhou.
\newblock Unicoder: A universal language encoder by pre-training with multiple
  cross-lingual tasks.
\newblock {\em arXiv preprint arXiv:1909.00964}, 2019.

\bibitem{huang2020pixel}
Zhicheng Huang, Zhaoyang Zeng, Bei Liu, Dongmei Fu, and Jianlong Fu.
\newblock Pixel-bert: Aligning image pixels with text by deep multi-modal
  transformers.
\newblock {\em arXiv preprint arXiv:2004.00849}, 2020.

\bibitem{Jiang2020HowCW}
Zhengbao Jiang, Frank~F. Xu, J. Araki, and Graham Neubig.
\newblock How can we know what language models know?
\newblock {\em TACL}, 8:423--438, 2020.

\bibitem{yfcc}
Sebastian Kalkowski, Christian Schulze, Andreas Dengel, and Damian Borth.
\newblock Real-time analysis and visualization of the yfcc100m dataset.
\newblock In {\em Proceedings of the 2015 workshop on community-organized
  multimodal mining: opportunities for novel solutions}, pages 25--30, 2015.

\bibitem{kay2017kinetics}
Will Kay, Joao Carreira, Karen Simonyan, Brian Zhang, Chloe Hillier, Sudheendra
  Vijayanarasimhan, Fabio Viola, Tim Green, Trevor Back, Paul Natsev, et~al.
\newblock The kinetics human action video dataset.
\newblock {\em arXiv preprint arXiv:1705.06950}, 2017.

\bibitem{kim2021vilt}
Wonjae Kim, Bokyung Son, and Ildoo Kim.
\newblock Vilt: Vision-and-language transformer without convolution or region
  supervision.
\newblock {\em arXiv preprint arXiv:2102.03334}, 2021.

\bibitem{kingma2014adam}
Diederik~P Kingma and Jimmy Ba.
\newblock Adam: A method for stochastic optimization.
\newblock {\em arXiv preprint arXiv:1412.6980}, 2014.

\bibitem{krishna2017visual}
Ranjay Krishna, Yuke Zhu, Oliver Groth, Justin Johnson, Kenji Hata, Joshua
  Kravitz, Stephanie Chen, Yannis Kalantidis, Li-Jia Li, David~A Shamma, et~al.
\newblock Visual genome: Connecting language and vision using crowdsourced
  dense image annotations.
\newblock {\em IJCV}, 123(1):32--73, 2017.

\bibitem{lan2019albert}
Zhenzhong Lan, Mingda Chen, Sebastian Goodman, Kevin Gimpel, Piyush Sharma, and
  Radu Soricut.
\newblock Albert: A lite bert for self-supervised learning of language
  representations.
\newblock {\em arXiv preprint arXiv:1909.11942}, 2019.

\bibitem{Larsson2017FractalNetUN}
Gustav Larsson, Michael Maire, and Gregory Shakhnarovich.
\newblock Fractalnet: Ultra-deep neural networks without residuals.
\newblock {\em arXiv preprint arXiv:1605.07648}, 2017.

\bibitem{Lester2021ThePO}
Brian Lester, Rami Al-Rfou, and Noah Constant.
\newblock The power of scale for parameter-efficient prompt tuning.
\newblock In {\em EMNLP}, 2021.

\bibitem{lewis2019bart}
Mike Lewis, Yinhan Liu, Naman Goyal, Marjan Ghazvininejad, Abdelrahman Mohamed,
  Omer Levy, Ves Stoyanov, and Luke Zettlemoyer.
\newblock Bart: Denoising sequence-to-sequence pre-training for natural
  language generation, translation, and comprehension.
\newblock {\em arXiv preprint arXiv:1910.13461}, 2019.

\bibitem{lewis2021paq}
Patrick Lewis, Yuxiang Wu, Linqing Liu, Pasquale Minervini, Heinrich Küttler,
  Aleksandra Piktus, Pontus Stenetorp, and Sebastian Riedel.
\newblock Paq: 65 million probably-asked questions and what you can do with
  them.
\newblock {\em arXiv preprint arXiv:2102.07033}, 2021.

\bibitem{Li2020UnicoderVLAU}
Gen Li, Nan Duan, Yuejian Fang, Daxin Jiang, and Ming Zhou.
\newblock Unicoder-vl: A universal encoder for vision and language by
  cross-modal pre-training.
\newblock In {\em AAAI}, 2020.

\bibitem{li2019visualbert}
Liunian~Harold Li, Mark Yatskar, Da Yin, Cho-Jui Hsieh, and Kai-Wei Chang.
\newblock Visualbert: A simple and performant baseline for vision and language.
\newblock {\em arXiv preprint arXiv:1908.03557}, 2019.

\bibitem{li2021prefixtuning}
Xiang~Lisa Li and Percy Liang.
\newblock Prefix-tuning: Optimizing continuous prompts for generation, 2021.

\bibitem{CNNSURVEY}
Zewen Li, Fan Liu, Wenjie Yang, Shouheng Peng, and Jun Zhou.
\newblock A survey of convolutional neural networks: Analysis, applications,
  and prospects.
\newblock {\em IEEE Transactions on Neural Networks and Learning Systems},
  pages 1--21, 2021.

\bibitem{lin2014microsoft}
Tsung-Yi Lin, Michael Maire, Serge Belongie, James Hays, Pietro Perona, Deva
  Ramanan, Piotr Doll{\'a}r, and C~Lawrence Zitnick.
\newblock Microsoft coco: Common objects in context.
\newblock In {\em ECCV}, 2014.

\bibitem{liu2021pre}
Pengfei Liu, Weizhe Yuan, Jinlan Fu, Zhengbao Jiang, Hiroaki Hayashi, and
  Graham Neubig.
\newblock Pre-train, prompt, and predict: A systematic survey of prompting
  methods in natural language processing.
\newblock {\em arXiv preprint arXiv:2107.13586}, 2021.

\bibitem{liuPtuningv2}
Xiao Liu, Kaixuan Ji, Yicheng Fu, Zhengxiao Du, Zhilin Yang, and Jie Tang.
\newblock P-tuning v2: Prompt tuning can be comparable to fine-tuning
  universally across scales and tasks.
\newblock {\em arXiv preprint arXiv:2110.07602}, 2021.

\bibitem{liu2021gpt}
Xiao Liu, Yanan Zheng, Zhengxiao Du, Ming Ding, Yujie Qian, Zhilin Yang, and
  Jie Tang.
\newblock Gpt understands, too.
\newblock {\em arXiv preprint arXiv:2103.10385}, 2021.

\bibitem{liu2019roberta}
Yinhan Liu, Myle Ott, Naman Goyal, Jingfei Du, Mandar Joshi, Danqi Chen, Omer
  Levy, Mike Lewis, Luke Zettlemoyer, and Veselin Stoyanov.
\newblock Roberta: A robustly optimized bert pretraining approach.
\newblock {\em arXiv preprint arXiv:1907.11692}, 2019.

\bibitem{liu2021Swin}
Ze Liu, Yutong Lin, Yue Cao, Han Hu, Yixuan Wei, Zheng Zhang, Stephen Lin, and
  Baining Guo.
\newblock Swin transformer: Hierarchical vision transformer using shifted
  windows.
\newblock {\em ICCV}, 2021.

\bibitem{Lu2019ViLBERTPT}
Jiasen Lu, Dhruv Batra, Devi Parikh, and Stefan Lee.
\newblock Vilbert: Pretraining task-agnostic visiolinguistic representations
  for vision-and-language tasks.
\newblock In {\em NeurIPS}, 2019.

\bibitem{lu2019vilbert}
Jiasen Lu, Dhruv Batra, Devi Parikh, and Stefan Lee.
\newblock Vilbert: Pretraining task-agnostic visiolinguistic representations
  for vision-and-language tasks.
\newblock {\em arXiv preprint arXiv:1908.02265}, 2019.

\bibitem{luo2021clip4clip}
Huaishao Luo, Lei Ji, Ming Zhong, Yang Chen, Wen Lei, Nan Duan, and Tianrui Li.
\newblock Clip4clip: An empirical study of clip for end to end video clip
  retrieval.
\newblock {\em arXiv preprint arXiv:2104.08860}, 2021.

\bibitem{monfort2019moments}
Mathew Monfort, Alex Andonian, Bolei Zhou, Kandan Ramakrishnan, Sarah~Adel
  Bargal, Tom Yan, Lisa Brown, Quanfu Fan, Dan Gutfreund, Carl Vondrick, et~al.
\newblock Moments in time dataset: one million videos for event understanding.
\newblock {\em TPAMI}, 42(2):502--508, 2019.

\bibitem{ordonez2011im2text}
Vicente Ordonez, Girish Kulkarni, and Tamara Berg.
\newblock Im2text: Describing images using 1 million captioned photographs.
\newblock {\em NeurIPS}, 24:1143--1151, 2011.

\bibitem{clipadapter}
Gao Peng, Geng Shijie, Zhang Renrui, Ma Teli, Fang Rongyao, Zhang Yongfeng, Li
  Hongsheng, and Qiao Yu.
\newblock Clip-adapter: Better vision-language models with feature adapters.
\newblock {\em arXiv preprint arXiv:2110.04544}, 2021.

\bibitem{plummer2015flickr30k}
Bryan~A Plummer, Liwei Wang, Chris~M Cervantes, Juan~C Caicedo, Julia
  Hockenmaier, and Svetlana Lazebnik.
\newblock Flickr30k entities: Collecting region-to-phrase correspondences for
  richer image-to-sentence models.
\newblock In {\em ICCV}, pages 2641--2649, 2015.

\bibitem{Qi2020ImageBERTCP}
Di Qi, Lin Su, Jianwei Song, Edward Cui, Taroon Bharti, and Arun Sacheti.
\newblock Imagebert: Cross-modal pre-training with large-scale weak-supervised
  image-text data.
\newblock {\em arXiv preprint arXiv:2001.07966}, 2020.

\bibitem{qiu2020pre}
Xipeng Qiu, Tianxiang Sun, Yige Xu, Yunfan Shao, Ning Dai, and Xuanjing Huang.
\newblock Pre-trained models for natural language processing: A survey.
\newblock {\em Science China Technological Sciences}, pages 1--26, 2020.

\bibitem{clip}
Alec Radford, Jong~Wook Kim, Chris Hallacy, Aditya Ramesh, Gabriel Goh,
  Sandhini Agarwal, Girish Sastry, Amanda Askell, Pamela Mishkin, Jack Clark,
  et~al.
\newblock Learning transferable visual models from natural language
  supervision.
\newblock {\em arXiv preprint arXiv:2103.00020}, 2021.

\bibitem{Radford2021LearningTV}
Alec Radford, Jong~Wook Kim, Chris Hallacy, Aditya Ramesh, Gabriel Goh,
  Sandhini Agarwal, Girish Sastry, Amanda Askell, Pamela Mishkin, Jack Clark,
  Gretchen Krueger, and Ilya Sutskever.
\newblock Learning transferable visual models from natural language
  supervision.
\newblock In {\em ICML}, 2021.

\bibitem{ren2015faster}
Shaoqing Ren, Kaiming He, Ross Girshick, and Jian Sun.
\newblock Faster r-cnn: Towards real-time object detection with region proposal
  networks.
\newblock {\em NeurIPS}, 28:91--99, 2015.

\bibitem{tipadapter}
Zhang Renrui, Fang Rongyao, Gao Peng, Zhang Wei, Li Kunchang, Dai Jifeng, Qiao
  Yu, and Li Hongsheng.
\newblock Tip-adapter: Training-free clip-adapter for better vision-language
  modeling.
\newblock {\em arXiv preprint arXiv:2111.03930}, 2021.

\bibitem{sennrich2016neural}
Rico Sennrich, Barry Haddow, and Alexandra Birch.
\newblock Neural machine translation of rare words with subword units.
\newblock In {\em ACL}, 2016.

\bibitem{sharma2018conceptual}
Piyush Sharma, Nan Ding, Sebastian Goodman, and Radu Soricut.
\newblock Conceptual captions: A cleaned, hypernymed, image alt-text dataset
  for automatic image captioning.
\newblock In {\em ACL}, pages 2556--2565, 2018.

\bibitem{autoprompt:emnlp20}
Taylor Shin, Yasaman Razeghi, Robert L.~Logan IV, Eric Wallace, and Sameer
  Singh.
\newblock {AutoPrompt}: Eliciting knowledge from language models with
  automatically generated prompts.
\newblock In {\em EMNLP}, 2020.

\bibitem{su2019vl}
Weijie Su, Xizhou Zhu, Yue Cao, Bin Li, Lewei Lu, Furu Wei, and Jifeng Dai.
\newblock Vl-bert: Pre-training of generic visual-linguistic representations.
\newblock {\em arXiv preprint arXiv:1908.08530}, 2019.

\bibitem{cbt}
Chen Sun, Fabien Baradel, Kevin Murphy, and Cordelia Schmid.
\newblock Learning video representations using contrastive bidirectional
  transformer.
\newblock {\em arXiv preprint arXiv:1906.05743}, 2019.

\bibitem{sun2019videobert}
Chen Sun, Austin Myers, Carl Vondrick, Kevin Murphy, and Cordelia Schmid.
\newblock Videobert: A joint model for video and language representation
  learning.
\newblock In {\em CVPR}, pages 7464--7473, 2019.

\bibitem{tan2019lxmert}
Hao Tan and Mohit Bansal.
\newblock Lxmert: Learning cross-modality encoder representations from
  transformers.
\newblock {\em arXiv preprint arXiv:1908.07490}, 2019.

\bibitem{pmlr-v139-touvron21a}
Hugo Touvron, Matthieu Cord, Matthijs Douze, Francisco Massa, Alexandre
  Sablayrolles, and Herve Jegou.
\newblock Training data-efficient image transformers \& distillation through
  attention.
\newblock In {\em ICML}, volume 139, pages 10347--10357, July 2021.

\bibitem{touvron2021training}
Hugo Touvron, Matthieu Cord, Matthijs Douze, Francisco Massa, Alexandre
  Sablayrolles, and Herv{\'e} J{\'e}gou.
\newblock Training data-efficient image transformers \& distillation through
  attention.
\newblock In {\em ICML}, pages 10347--10357. PMLR, 2021.

\bibitem{touvron2021cait}
Hugo Touvron, Matthieu Cord, Alexandre Sablayrolles, Gabriel Synnaeve, and
  Herv\'e J\'egou.
\newblock Going deeper with image transformers.
\newblock {\em arXiv preprint arXiv:2103.17239}, 2021.

\bibitem{vaswani2017attention}
Ashish Vaswani, Noam Shazeer, Niki Parmar, Jakob Uszkoreit, Llion Jones,
  Aidan~N Gomez, {\L}ukasz Kaiser, and Illia Polosukhin.
\newblock Attention is all you need.
\newblock In {\em NeurIPS}, pages 5998--6008, 2017.

\bibitem{wang2018glue}
Alex Wang, Amanpreet Singh, Julian Michael, Felix Hill, Omer Levy, and Samuel~R
  Bowman.
\newblock Glue: A multi-task benchmark and analysis platform for natural
  language understanding.
\newblock {\em arXiv preprint arXiv:1804.07461}, 2018.

\bibitem{wang2021pyramid}
Wenhai Wang, Enze Xie, Xiang Li, Deng-Ping Fan, Kaitao Song, Ding Liang, Tong
  Lu, Ping Luo, and Ling Shao.
\newblock Pyramid vision transformer: A versatile backbone for dense prediction
  without convolutions.
\newblock In {\em ICCV}, 2021.

\bibitem{xu2021pre}
Han Xu, Zhang Zhengyan, Ding Ning, Gu Yuxian, Liu Xiao, Huo Yuqi, Qiu Jiezhong,
  Zhang Liang, Han Wentao, Huang Minlie, et~al.
\newblock Pre-trained models: Past, present and future.
\newblock {\em arXiv preprint arXiv:2106.07139}, 2021.

\bibitem{Yuan_2021_ICCV}
Li Yuan, Yunpeng Chen, Tao Wang, Weihao Yu, Yujun Shi, Zi-Hang Jiang,
  Francis~E.H. Tay, Jiashi Feng, and Shuicheng Yan.
\newblock Tokens-to-token vit: Training vision transformers from scratch on
  imagenet.
\newblock In {\em ICCV}, pages 558--567, October 2021.

\bibitem{Zhang_2021_ICCV}
Yanyi Zhang, Xinyu Li, Chunhui Liu, Bing Shuai, Yi Zhu, Biagio Brattoli, Hao
  Chen, Ivan Marsic, and Joseph Tighe.
\newblock Vidtr: Video transformer without convolutions.
\newblock In {\em ICCV}, pages 13577--13587, October 2021.

\bibitem{zhang2020object}
Ziqi Zhang, Yaya Shi, Chunfeng Yuan, Bing Li, Peijin Wang, Weiming Hu, and
  Zheng-Jun Zha.
\newblock Object relational graph with teacher-recommended learning for video
  captioning.
\newblock In {\em CVPR}, pages 13278--13288, 2020.

\bibitem{zhou2021deepvit}
Daquan Zhou, Bingyi Kang, Xiaojie Jin, Linjie Yang, Xiaochen Lian, Qibin Hou,
  and Jiashi Feng.
\newblock Deepvit: Towards deeper vision transformer.
\newblock {\em arXiv preprint arXiv:2103.11886}, 2021.

\bibitem{Zhou2021LearningTP}
Kaiyang Zhou, Jingkang Yang, Chen~Change Loy, and Ziwei Liu.
\newblock Learning to prompt for vision-language models.
\newblock {\em arXiv preprint arXiv:2109.01134}, 2021.

\bibitem{zhou2020unified}
Luowei Zhou, Hamid Palangi, Lei Zhang, Houdong Hu, Jason Corso, and Jianfeng
  Gao.
\newblock Unified vision-language pre-training for image captioning and vqa.
\newblock In {\em AAAI}, 2020.

\bibitem{zhu2015aligning}
Yukun Zhu, Ryan Kiros, Rich Zemel, Ruslan Salakhutdinov, Raquel Urtasun,
  Antonio Torralba, and Sanja Fidler.
\newblock Aligning books and movies: Towards story-like visual explanations by
  watching movies and reading books.
\newblock In {\em ICCV}, pages 19--27, 2015.

\end{thebibliography}
}

\clearpage
\appendix
\section{Tokenizer}
Given the raw inputs from text, image, and video modalities, modality-specific tokenizers are applied to generate the input token sequences for the Transformer encoder.
Here, we use the BPE tokenizer~\cite{sennrich2016neural} for text modality, the image patch tokenizer~\cite{dosovitskiy2020image} for image modality, and the temporal frame patch tokenizer~\cite{bertasius2021space} for video modality. These outputted tokens are attached with additional modality type embeddings to identify which modality the raw input belongs to. The tokenizers are illustrated in Fig.~\ref{fig:overview}.

\vspace{0.5em}\noindent\textbf{Text Tokenizer.~} The BPE Tokenizer~\cite{sennrich2016neural} is employed for text modality. The text inputs are split into sub-words and projected by a linear embedding layer. A learnable 1D positional embedding for text with a max length of 256 is added to the word embeddings. To specify the input modality, an additional trainable textual modality embedding \verb|<T>| is added to each token embedding. Note that all the text inputs of our model share the same vocabulary.

\vspace{0.5em}\noindent\textbf{Image Tokenizer.~} The image patch tokenizer~\cite{dosovitskiy2020image} is utilized as the image tokenizer. The input images are resized to 224$\times$224, and flattened to a sequence of image patches with shapes 16$\times$16, which are further mapped by a linear projection. A sequence of learnable image positional embeddings with a fixed length $14\times14 = 196$, and an additional visual modality embedding \verb|<V>| are also added.

\vspace{0.5em}\noindent\textbf{Video Tokenizer.~} The temporal frame patch tokenizer~\cite{bertasius2021space} is used for video tokenization. Each frame of the input video is flattened to image patches with shape 16$\times$16. For a video with $N$ frames ($N = 8$ by default), the number of tokens would be $N\times14\times14$. The spatial positional embeddings for images as well as a 1D temporal position embedding with max length $N=8$ are added to the video embeddings. Besides, the additional visual modality embedding \verb|<V>| is added to each token embedding.

\section{Implementation Details for Auto-regressive Language Modeling}
\begin{figure}[t]
\centering
\includegraphics[width=0.45\textwidth]{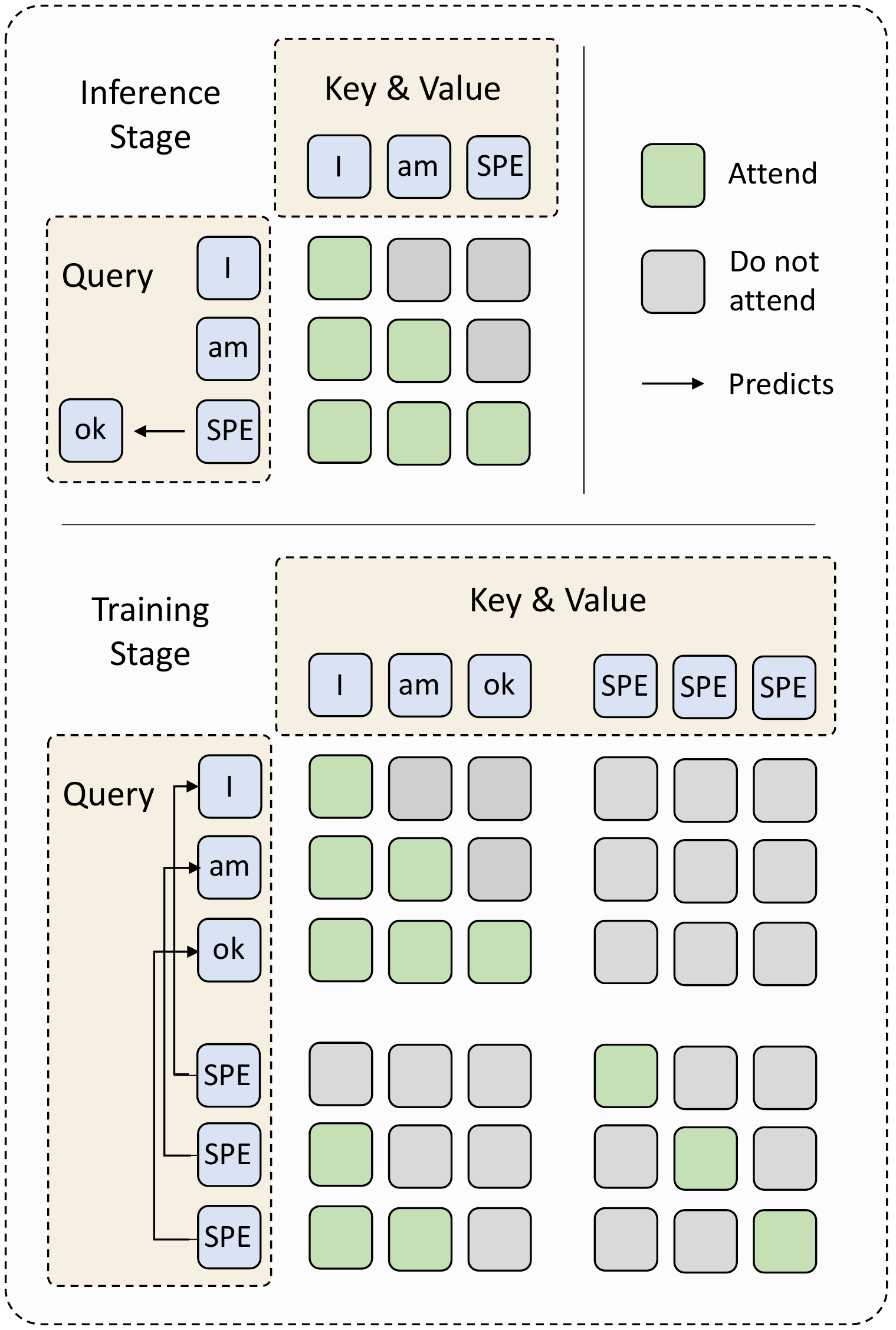}
\caption{The attention mask of autoregressive language modeling.}
\label{fig:autoregressive}
\end{figure}
In the formulation of autoregressive tasks from previous works, each input token attends to all previous tokens, including itself, to predict \textbf{the next word}. 
Unlike previous works, we use \verb|<SPE>| to predict \textbf{the current word}. 
Fig.~\ref{fig:autoregressive} shows how we achieve this. In the inference stage, \texttt{<SPE>} is appended to the end of the predicted tokens as the input. The corresponding output is used for prediction. 
In the training stage, we append several \texttt{<SPE>} tokens after the input sequence. Each \texttt{<SPE>} token is trained to predict a word in the input sequence. 
The attention mask is designed to make sure that the word tokens do not attend to \texttt{<SPE>} tokens, and each \texttt{<SPE>} token only attends to itself and the previous word tokens. In this way, the training stage and the inference stage are aligned.

\label{impleofauto-regre}
\section{Prompt Tuning}
Four groups of parameters are learnable in prompt-tuning. They are \verb|<SPE>|, layer normalization parameters, prompt tokens, and linear heads. This section introduces the usage and the number of parameters of each parameter group.

\vspace{0.4em}\noindent\textbf{Details} Similar to the pre-training stage, \verb|<SPE>| token is shared for both inputs and targets.
Layer normalization weights and biases in each Transformer layer and tokenizer are tuned. 
Learnable prompts are added to each layer of the Transformer encoder. 
Specifically, the input prompts of each layer do not come from the output of the previous layer. 
They are random initialized learnable parameters. 
For all prompt-tuning experiments, we use 10 learnable prompts for each layer on both inputs.
The linear heads only apply for classification tasks.
It takes the feature that is to be classified as input, and returns a classification logit.
The output probability is a linear combination of the probability from the similarity score and the linear head:
\begin{equation}
\label{eq:pt}
P(x, y) \propto \alpha \exp\left(\cos\big(f(x), f(y)\big) / \tau \right) + w^\top f(x) + b
\end{equation}
where $w$ and $b$ are the weights and bias of the linear, and $\alpha$ is a learnable scalar. 
$w$ and $b$ are initialized with 0 and $\alpha$ is initialized with 1.

\vspace{0.4em}\noindent\textbf{Number of Parameters} 
\begin{table} [t]
    \centering
\resizebox{0.33\textwidth}{!}{
    \begin{tabular}{P{5em} P{10em} }
    \toprule
      & Number of Parameters  \\
  \midrule
      \texttt{<SPE>} & 768 \\
      Layer Norm & 41,472 \\
      Prompt & 184,320 \\
      Linear head & 768 * \texttt{num\_classes} \\
   \bottomrule   
    \end{tabular}
}
    \caption{The number of learnable parameters in the prompt-tuning stage. Note that the linear head only applies for classification tasks.}
    \label{tab:ptuning_stat}
    \vspace{-0.6em}
\end{table}
Tab.~\ref{tab:ptuning_stat} shows the number of parameters of each learnable component in prompt-tuning. 
For tasks other than classification, the number of parameters is 227K in total. 
When the linear head is added, \eg, in ImageNet-1k classification task, the number is 995K, which is still less than 1\% of all the parameters in the pre-training stage.

\section{Formulation of Novel Tasks}

\begin{figure}[t]
\centering
\includegraphics[width=0.49\textwidth]{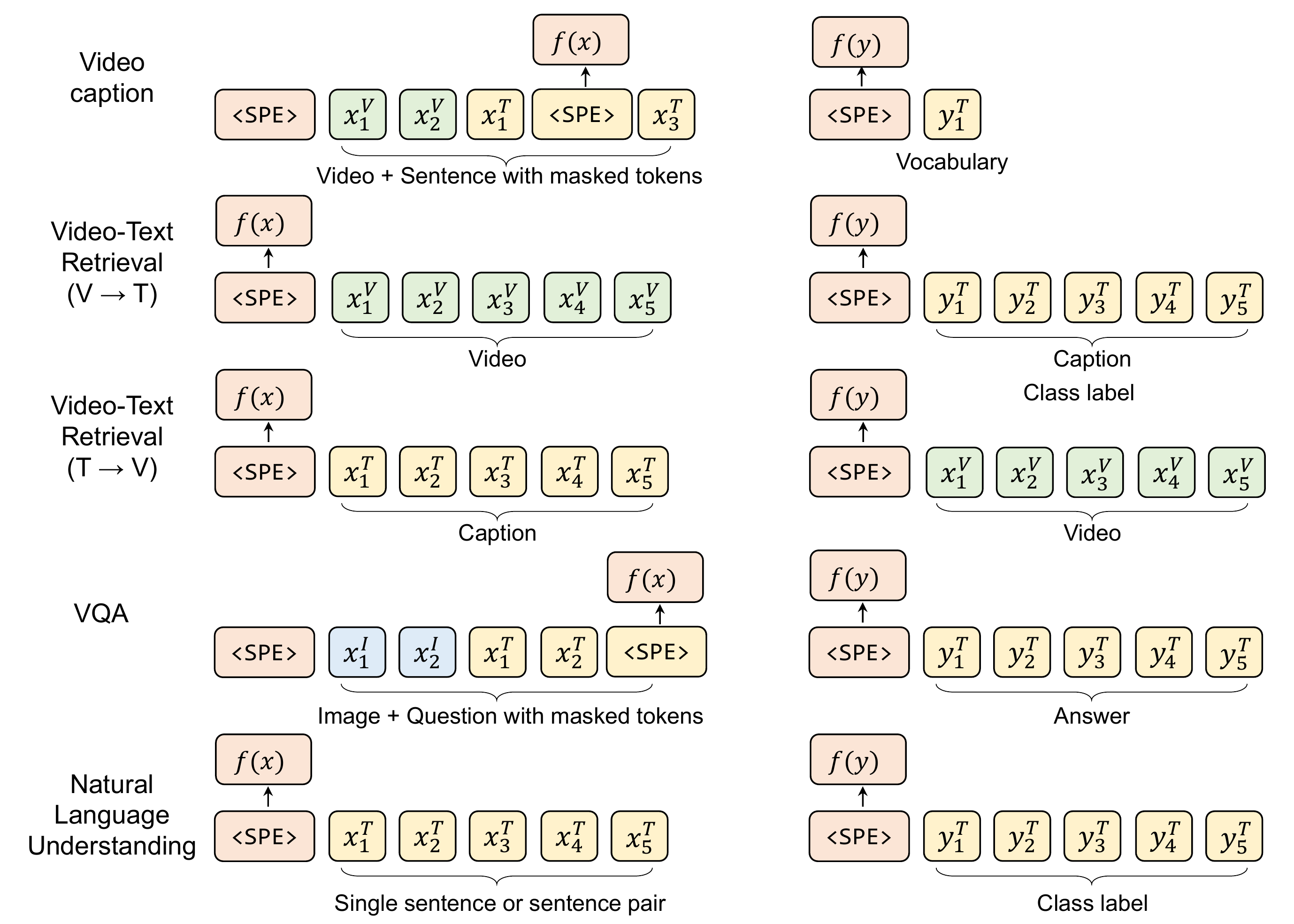}
\caption{Input and target formats of our novel tasks. For each task, the left column represents the format of input sequence $x$, and the right column represents the format of the target sequence $y$. $f(x)$ and $f(y)$ are used to calculate the joint probability distribution. Here, we have omitted the tokenizer and encoder for concision.}
\label{fig:novel_tasks}
\vspace{-1em}
\end{figure}
The generic perceptual modeling makes it  easy to convert existing tasks into the unified task formulation of our Uni-Perceiver. 
Fig.~\ref{fig:novel_tasks} illustrates the input and  the output formulations of our novel tasks, which  we will describe  in details.

\vspace{0.4em}\noindent\textbf{Video Caption.~}
Similar to image caption, video caption is modeled as  the autoregressive  language modeling task with video clues.
In this task, the model predicts each word based on the video and its previous words. 
The input set $\mathcal{X}$ consists of the sequence concatenation of  video and the words that have been predicted, followed with a  \verb|<SPE>| token.
The output features of the  \verb|<SPE>| token is used to calculate the joint probability with each words in the  the vocabulary  set $\mathcal{Y}$.
Then the word with the highest probability is the predicted word at the current location.
Additionally, video caption follows the efficient implementation of autoregressive training  introduced in Sec.~\ref{impleofauto-regre}.

\vspace{0.4em}\noindent\textbf{Video-Text Retrieval.~}
The Video-Text retrieval follows the formulation of Image-Text retrieval task, except that the image sequence is replaced by the video sequence.
Specifically, the input sets  $\mathcal{X}$ and  $\mathcal{Y}$ are composed of video and text sequences respectively.
Each sequence in  $\mathcal{X}$ and  $\mathcal{Y}$  also has a \verb|<SPE>| token at the beginning.
We use the output feature at the \verb|<SPE>| token as the final representation of the input video or the text to calculate the joint probability distribution.

\vspace{0.4em}\noindent\textbf{Visual Question Answering.~} 
We formulate the VQA task as a special case of masked language modeling with image clues. 
The input $x \in \mathcal{X}$ is the combination of image and question sequences. It should be noted that each question ends with a  ``?'' mark  and  a \verb|<SPE>|  token is appended to the end of the question sequence.
The $\mathcal{Y}$ set consists of candidate answer sequences, in which each begins with a \verb|<SPE>|  token too.
The joint probability distribution between $\mathcal{X}$ and $\mathcal{Y}$ can be calculated by using the output feature from \verb|<SPE>|  tokens.

\vspace{0.4em}\noindent\textbf{Natural Language Understanding.~}
The formulation of natural language understanding task is similar to that of image classification task. 
 $\mathcal{X}$ denotes the set of the input single sentence or the sentence-pair, and   $\mathcal{Y}$ is the set contains the textual class label. 
 For example, in SST-2 \citeappendix{SST2},  $x$ denotes the sentence sequence of  movie review and  $y \in \mathcal{Y} = \{\verb|great|, \verb|terrible|\}$ is the sentiment label.
 In MRPC \citeappendix{mrpc},  $x $  instead  is the sequence  combination of sentence pairs extracted from news, and  $y \in \mathcal{Y} = \{\verb|Yes|, \verb|No|\}$ is the label    to indicate whether the sentence pair are semantically equivalent.
 We also add \verb|<SPE>| tokens at the beginning of the sequences $x$ and $y$, of which output features are used to computed joint probability.

\section{Extra pre-training details}

\begin{table}[t]
\small
    \centering
\resizebox{0.48\textwidth}{!}{
\begin{tabular}{c|ccc}

\toprule
Task & Dataset & Batch Size & Sampling Weight  \\

\midrule
\multirow{1}{*} Image Classification & ImageNet-21k~\cite{deng2009imagenet} & 64 &  0.333\\
\midrule
\multirow{2}{*} {Video Classification} & Kinetics-700~\cite{kay2017kinetics} & 4 &  0.0925\\
                                        & Moments in Time~\cite{monfort2019moments} & 24 & 0.0185 \\
\midrule
\multirow{8}{*} {Auto-encoding LM} & Books\&Wiki~\cite{zhu2015aligning} & 64 &  0.07775\\
                                        & YFCC~\cite{yfcc} & 64 & 0.02778 \\
                                        & CC12M~\cite{changpinyo2021cc12m} & 64 & 0.02778 \\
                                        & CC3M~\cite{sharma2018conceptual} & 64 & 0.01389 \\
                                        & Visual Genome~\cite{krishna2017visual} & 64 & 0.01389 \\
                                        & COCO Caption~\cite{Chen2015MicrosoftCC} & 64 & 0.01389 \\
                                        & SBU~\cite{ordonez2011im2text} & 64 & 0.01389 \\
                                        & PAQ~\cite{lewis2021paq} & 512 & 0.0222 \\
\midrule
\multirow{8}{*} {Auto-regressive LM} & Books\&Wiki~\cite{zhu2015aligning} & 64 &  0.07775\\
                                        & YFCC~\cite{yfcc} & 56 & 0.02778 \\
                                        & CC12M~\cite{changpinyo2021cc12m} & 56 & 0.02778 \\
                                        & CC3M~\cite{sharma2018conceptual} & 56 & 0.01389 \\
                                        & Visual Genome~\cite{krishna2017visual} & 56 & 0.01389 \\
                                        & COCO Caption~\cite{Chen2015MicrosoftCC} & 56 & 0.01389 \\
                                        & SBU~\cite{ordonez2011im2text} & 56 & 0.01389 \\
                                        & PAQ~\cite{lewis2021paq} & 400 & 0.0222 \\
\midrule
\multirow{7}{*} {Retrieval} 
                                        & YFCC~\cite{yfcc} & 128 & 0.02778 \\
                                        & CC12M~\cite{changpinyo2021cc12m} & 128 & 0.02778 \\
                                        & CC3M~\cite{sharma2018conceptual} & 128 & 0.01389 \\
                                        & Visual Genome~\cite{krishna2017visual} & 128 & 0.01389 \\
                                        & COCO Caption~\cite{Chen2015MicrosoftCC} & 128 & 0.01389 \\
                                        & SBU~\cite{ordonez2011im2text} & 128 & 0.01389 \\
                                        & PAQ~\cite{lewis2021paq} & 512 & 0.0222 \\

\bottomrule
\end{tabular}}
    
    \caption{
    Ingredients and hyper-parameters for our pre-training.
    } 
    \label{tab:trianingdetails}
    \vspace{-1.2em}
\end{table}
\vspace{0.4em}\noindent\textbf{Sampling Weight \& Batch Size \& Loss ~} Tab.~\ref{tab:trianingdetails} lists the batch size and sampling weight of each task and dataset in the pre-training stage. 
We use cross-entropy loss for language modeling tasks. The other tasks are trained with cross-entropy loss with 0.1 label smoothing. 
The loss weight of video classification is 0.05, which helps stabilize training in our experiments. The other loss weights are 1.0 by default.

For retrieval tasks like image-text retrieval, we use training  samples in the same batch as negative samples, whose typical size is 127 except the PAQ dataset.
Note that we do not use memory bank and do not  gather feature across GPU devices to provide more negative samples, which may  further promote the performance of retrieval tasks.

\vspace{0.4em}\noindent\textbf{Data Augmentation ~}
We apply augmentation techniques to image and video modalities to avoid overfitting.
For images in ImageNet-21k dataset, we apply augmentation same as \cite{touvron2021training}.
Rand-Aug\citeappendix{cubuk2020randaugment}, random erasing \citeappendix{zhong2020random}, mixup \citeappendix{zhang2017mixup} and cutmix \citeappendix{yun2019cutmix} are  used simultaneously.
For images in other datasets, we resize the images to the  short edge size of 256, and then   a $224\times224$  region is cropped randomly from  the resized images during training.
During inference, the random crop is replaced with center crop operation.
For all Video inputs, we apply the same augmentations used in \cite{bertasius2021space}. We use clips of size $8\times224\times224$ for Kinetics-700 and Kinetics-400, and $3\times224\times224$ for Moments in Time. The temporal sample rate is 32. We use a single temporal clip. During training, the start frame is randomly picked if the video is longer than the clip.
In the training stage, we first resize the shorter side of the video to a random value in $[256,320]$, then we randomly sample a $224\times224$ crop. In the test stage, the short side of the video is resized to 224. For classification tasks, We use 3 spatial crops with size $224\times224$ to cover a larger range of content and average their logits for evaluation. For video captioning task, we use center crop after resizing.

\vspace{0.4em}\noindent\textbf{Data Parallel for Vocabulary and Class Labels ~}
Naive implementation of language modeling and ImageNet-21k classification is impractical due to memory limitation. In language modeling, we need to compare the feature of a \texttt{<SPE>} token in a sequence to the feature of each token in our tokenizer. Since the vocabulary size is large, we apply data parallel for the vocabulary set. A similar method is used for class labels of ImageNet-21k.

\vspace{0.4em}\noindent\textbf{Removing Overlap ~}
For the training set of K700 participating in pre-training, we remove those videos  overlapping with validation set of K400.

\section{Licences of Datasets}

\vspace{0.4em}\noindent\textbf{ImageNet-21K} \cite{deng2009imagenet} is subject to the ImageNet terms of use \citeappendix{imagenetterms}.

\vspace{0.4em}\noindent\textbf{Kinetics-700} \citeappendix{Smaira2020ASN} \& \textbf{Kinetics-400} \cite{kay2017kinetics} The kinetics dataset is licensed by Google Inc. under a Creative Commons Attribution 4.0 International License.

\vspace{0.4em}\noindent\textbf{BooksCorpus} \cite{zhu2015aligning} Replicate Toronto BookCorpus is open-source and licensed under GNU GPL, Version 3.

\vspace{0.4em}\noindent\textbf{Wikipedia}  Most of Wikipedia's text is co-licensed under the Creative Commons Attribution-ShareAlike 3.0 Unported License (CC BY-SA) and the GNU Free Documentation License (GFDL) (unversioned, with no invariant sections, front-cover texts, or back-cover texts). Some text has been imported only under CC BY-SA and CC BY-SA-compatible license and cannot be reused under GFDL.

\vspace{0.4em}\noindent\textbf{YFCC} \cite{yfcc} All the photos and videos provided in YFCC dataset are licensed under one of the Creative Commons copyright licenses.

 \vspace{0.4em}\noindent\textbf{CC12M}~\cite{changpinyo2021cc12m} is licensed under  the Terms of Use of Conceptual 12M \citeappendix{cc12mlicense}.
 
 \vspace{0.4em}\noindent \textbf{CC3M}~\cite{sharma2018conceptual} is licensed under  the Conceptual Captions  Terms of Use  \citeappendix{cc3mlicense}.
  
 \vspace{0.4em}\noindent \textbf{Visual Genome}~\cite{krishna2017visual} is licensed under a Creative Commons Attribution 4.0 International License \citeappendix{vgterms}.
  
\vspace{0.4em}\noindent  \textbf{COCO Caption}~\cite{Chen2015MicrosoftCC} The images are subject to the Flickr terms of use~\citeappendix{flickr2020terms}.

 \vspace{0.4em}\noindent  \textbf{SBU Caption}~\cite{ordonez2011im2text} The images are subject to the Flickr terms of use~\citeappendix{flickr2020terms}.
   
 \vspace{0.4em}\noindent  \textbf{PAQ}~\cite{lewis2021paq} is licensed under the Attribution-NonCommercial 4.0 International License.

\vspace{2em}
{\small
\bibliographystyleappendix{ieee_fullname}
\bibliographyappendix{egbib}
}

\end{document}